%% file: main.tex
\documentclass[10pt,twocolumn,letterpaper]{article}

\usepackage[pagenumbers]{wacv} 
\makeatletter
\@namedef{ver@everyshi.sty}{}
\makeatother
\usepackage{tikz}

\usepackage{graphicx}
\usepackage{amsmath}
\usepackage{amssymb}
\usepackage{booktabs}

\usepackage{times}
\usepackage{epsfig}
\usepackage{graphicx}
\PassOptionsToPackage{usenames,dvipsnames}{xcolor}

\usepackage{booktabs}
\usepackage{soul}
\usepackage{enumitem}

\usepackage[pagebackref,breaklinks,colorlinks]{hyperref}

\usepackage[capitalize]{cleveref}
\crefname{section}{Sec.}{Secs.}
\Crefname{section}{Section}{Sections}
\Crefname{table}{Table}{Tables}
\crefname{table}{Tab.}{Tabs.}

\input{fig/fig_preamble}
\begin{document}

\input{src/macro}

\title{UVCGAN v2: An Improved Cycle-Consistent GAN for Unpaired Image-to-Image Translation}

\input{src/authors}

\maketitle

\input{src/abstract}
\input{src/intro}
\input{src/related_work}
\input{src/method}

\input{src/experiments}
\input{src/results}

\input{src/conclusions}

{\small
\bibliographystyle{ieee_fullname}
\bibliography{refs}
}

\newpage
\appendix

\input{src/app_data}
\input{src/app_train}
\input{src/app_eval_inconsistency}
\input{src/app_fidelity}
\input{src/app_results_consistent}
\input{table/app_results_consitency_effect}

\input{table/app_uvcgan_consist}

\input{src/app_figures}
\input{src/app_consistency_loss}

\input{src/app_uvcgan}

\end{document}

%% file: fig/fig_preamble.tex
\usepackage{tikz}
\usepackage{pgfplots}
\usepackage{ifthen}
\usetikzlibrary{
    fit,
    math,
    calc,
    arrows,
    shapes,
    shadows,
    fadings,
    external,
    plotmarks,
    arrows.meta,
    backgrounds,
    positioning,
    shadows.blur,
    intersections,
    pgfplots.groupplots,
    pgfplots.statistics,
    pgfplots.fillbetween,
}

\definecolor{b1}{HTML}{099CF0}
\definecolor{b2}{HTML}{0FA5C7}
\definecolor{yg}{HTML}{A1CB2E}
\definecolor{pg}{HTML}{5AAA4F}
\definecolor{or}{HTML}{F8A300}
\definecolor{pr}{HTML}{7000AD}
\definecolor{gr}{HTML}{EAEAF2}

\pgfdeclarelayer{back}
\pgfsetlayers{back,main}

\makeatletter
\pgfkeys{%
  /tikz/on layer/.code={
    \def\tikz@path@do@at@end{\endpgfonlayer\endgroup\tikz@path@do@at@end}%
    \pgfonlayer{#1}\begingroup%
  }%
}
\makeatother

\newcommand{\round}[2][2]{
    \pgfmathparse{#2}
    \pgfmathprintnumber[precision=#1, zerofill]{\pgfmathresult}
}

\makeatletter
\pgfplotsset{
    boxplot/hide outliers/.code={
        \def\pgfplotsplothandlerboxplot@outlier{}%
    }
}
\makeatother

%% file: src/macro.tex
\newcommand{\thename}{UVCGANv2\xspace}

\newcommand{\uvcgan}{UVCGAN\xspace}
\newcommand{\stylegan}{StyleGAN\xspace}
\newcommand{\styleganTwo}{StyleGANv2\xspace}
\newcommand{\cyclegan}{CycleGAN\xspace}
\newcommand{\aclgan}{ACLGAN\xspace}
\newcommand{\council}{CouncilGAN\xspace}
\newcommand{\progan}{ProGAN\xspace}
\newcommand{\ugatit}{U-GAT-IT\xspace}
\newcommand{\cut}{CUT\xspace}
\newcommand{\egsde}{EGSDE\xspace}
\newcommand{\egsdeDG}{EGSDE$^\dagger$\xspace}
\newcommand{\ittr}{ITTR\xspace}
\newcommand{\somesim}{LSeSim\xspace}
\newcommand{\ilvr}{ILVR\xspace}
\newcommand{\sdedit}{SDEdit\xspace}

\newcommand{\patchgan}{PatchGAN\xspace}

\newcommand{\sota}{state-of-the-art\xspace}
\newcommand{\lin}[1]{\textbf{\color{red}#1\xspace}}

\newcommand{\dmitrii}[1]{\textbf{\color{blue}#1\xspace}}
\newcommand{\yi}[1]{\textbf{\color{teal}#1\xspace}}
\newcommand{\ray}[1]{\textbf{\color{olive}#1\xspace}}

\newcommand{\domA}{$a$\xspace}
\newcommand{\domB}{$b$\xspace}

\newcommand{\anime}{Anime\xspace}
\newcommand{\celeba}{CelebA\xspace}
\newcommand{\celebahq}{CelebA-HQ\xspace}
\newcommand{\afhq}{AFHQ\xspace}

\newcommand{\selfieanime}{Selfie-to-Anime\xspace}
\newcommand{\animeselfie}{Anime-to-Selfie\xspace}
\newcommand{\gender}{Male-to-Female\xspace}
\newcommand{\malefemale}{Male-to-Female\xspace}
\newcommand{\femalemale}{Female-to-Male\xspace}
\newcommand{\glasses}{Glasses\xspace}
\newcommand{\rmvGlasses}{Removing Glasses\xspace}
\newcommand{\addGlasses}{Adding Glasses\xspace}
\newcommand{\catdog}{Cat-to-Dog\xspace}
\newcommand{\wilddog}{Wild-to-Dog\xspace}
\newcommand{\wildcat}{Wild-to-Cat\xspace}

\newcommand{\paren}[1]{\left(#1\right)}
\newcommand{\set}[1]{\left\{#1\right\}}
\newcommand{\abs}[1]{\left|#1\right|}
\newcommand{\norm}[1]{\left\Vert#1\right\Vert}

\newcommand{\fake}[1]{#1_{\textrm{f}}}
\newcommand{\reco}[1]{#1_{\textrm{c}}}
\newcommand{\iden}[1]{#1_{\textrm{i}}}
\newcommand{\gab}{\mathcal{G}_{A \rightarrow B}}
\newcommand{\gba}{\mathcal{G}_{B \rightarrow A}}
\newcommand{\da}{\mathcal{D}_{A}}
\newcommand{\db}{\mathcal{D}_{B}}

\renewcommand{\paragraph}[1]{\textbf{#1}}

%% file: src/authors.tex
\author{
    Dmitrii Torbunov, Yi Huang, Huan-Hsin Tseng, Haiwang Yu, \\Jin Huang, Shinjae Yoo, Meifeng Lin, Brett Viren, Yihui Ren\\
    Brookhaven National Laboratory, Upton, NY, USA\\
    {\tt\small \{dtorbunov,yhuang2,htseng,hyu,jhuang,sjyoo,mlin,bviren,yren\}@bnl.gov}
}

%% file: src/abstract.tex
\begin{abstract}
An unpaired image-to-image (I2I) translation technique seeks to find a mapping between
two domains of data in a fully unsupervised manner.
While initial solutions to the I2I problem were provided by generative adversarial neural networks (GANs), diffusion models (DMs) currently hold the state-of-the-art status on the I2I translation benchmarks in terms of Fréchet inception distance (FID).
Yet, DMs suffer from limitations, such as not using data from the source domain during the training or maintaining consistency of the source and translated images
only via simple pixel-wise errors.
This work improves a recent \uvcgan model and equips it with modern advancements in model architectures and training procedures.
The resulting revised model significantly outperforms other advanced GAN- and DM-based competitors on a variety of benchmarks.
In the case of \malefemale translation of \celeba, the model achieves more than $40\%$ improvement in FID score compared to the \sota results.
This work also demonstrates the ineffectiveness of the pixel-wise I2I translation faithfulness metrics and suggests their revision.
The code and trained models are available at \url{https://github.com/LS4GAN/uvcgan2}.

\end{abstract}

%% file: src/intro.tex
\section{Introduction}

Image-to-image (I2I) translation models aim to find a mapping between two domains of images. When paired examples of images from two domains are available, such a mapping can be easily solved in a supervised manner. However, the unpaired I2I translation, where examples of pairs are not available, poses a more interesting problem. The ability to perform an unpaired I2I translation is highly beneficial as obtaining paired datasets in the real world is often difficult, time-consuming, or even impossible~\cite{boulanger2021deep}.

The advancement of unpaired I2I largely benefits from recent developments in deep generating models, such as (variational) Autoencoder~\cite{hinton2006reducing, kingma2013auto}, generative adversarial networks (GANs), and generating flows~\cite{rezende2015variational, dinh2014nice, dinh2016density, kingma2018glow}. One early successful unpaired I2I model is \cyclegan~\cite{zhu2017unpaired}, which uses a cycle-consistency constraint that requires a cyclic back-and-forth translation between two domains to produce the original image. Several succeeding models inspired by \cyclegan, such as STARGAN~\cite{choi2018stargan,choi2020stargan}, SEAN~\cite{zhu2020sean}, U-GAT-IT~\cite{kim2019u}, and CUT~\cite{park2020contrastive}, are designed to further enhance the quality and diversity of the generated images. However, GAN-based I2I methods lag behind general developments in GAN architecture and training procedures~\cite{karras2020analyzing,thanh2019improving}.

Meanwhile, \textit{diffusion models} (DMs) provide an alternative route to image generation~\cite{ho2020denoising}. With a recent spike of interest in such models, several
applications of DMs to unpaired I2I translation have been developed~\cite{choi2021ilvr,meng2021sdedit,zhao2022egsde}. Despite being relatively new, the DM-based \egsde~\cite{zhao2022egsde} approach already has demonstrated superior results on several benchmarks. However, because they do not use source images during the training, DM-based solutions may perform a suboptimal translation~\cite{zhao2022egsde}.
Additionally, DM-based methods rely on pixel-wise $L_2$ distances to maintain the consistency of the source and translated images.
This simple consistency measure is not guaranteed to preserve any semantically meaningful features and can restrict image transformations.

On multiple occasions, revisiting a classic neural architecture and improving it with
a number of modern additions has led to vast improvements in 
performance~\cite{liu2022convnet,karras2020analyzing,bello2021revisiting}.
Based on this observation, \uvcgan~\cite{torbunov2023uvcgan} modernizes the classic CycleGAN
architecture and achieves state-of-the-art performance on several I2I tasks.
Unlike the DM-based models, \cyclegan's training procedure is able to utilize
images from the source and target domains effectively and simultaneously.
Moreover, \cyclegan maintains an intrinsic consistency between the source and translated images (via the cycle-consistency constraint)---a feature that cannot be achieved by simple pixel-wise consistency measures.

Following the \uvcgan's spirit,
we introduce additional architectural innovations and bring modern training techniques to \cyclegan.
The resulting model significantly outperforms both \uvcgan (e.g., $50\%$ better realism on CelebA male-to-female)
and the state-of-the-art DM-based EGSDE (e.g., $40\%$ better on CelebA-HQ male-to-female). The following summarizes the contributions of this work:

\begin{itemize}[nosep,leftmargin=1em]
  \item We introduce a novel hybrid neural architecture for image translation,
    combining benefits of ViT and Style Modulated Convolutional blocks.
  \item We propose enhancing a common I2I discriminator architecture with a specialized head
    to prevent the problem of mode collapse~\cite{goodfellow2016nips}.
  \item We further improve the \uvcgan's training procedure by incorporating modern GAN training techniques
        that are not commonly applied to unpaired I2I problems.
  \item We perform extensive quantitative evaluations of our model and demonstrate it
    surpasses state-of-the-art competitors in terms of image realism on various translation tasks.
  \item Finally, we highlight inconsistencies and limitations of current unpaired I2I
    evaluation metrics and procedures. In response, we propose improved faithfulness measures that
    are better aligned with human perception.
\end{itemize}

%% file: src/related_work.tex
\section{Related Works}

Problems related to unpaired I2I translation have been approached from multiple
perspectives. There are two major classes of solutions: GAN- and diffusion-based methods.

\paragraph{GAN-based Methods.}
Multiple GAN-based methods have been developed to tackle the problem of unpaired I2I
translation. One distinct group of GAN-based methods involves methods that rely on cycle consistency,
including \cyclegan~\cite{zhu2017unpaired}, DualGAN~\cite{yi2017dualgan}, \ugatit~\cite{kim2019u},
and the recent \uvcgan~\cite{torbunov2023uvcgan}. This class of algorithms requires two generator networks that translate
images in opposite directions. Its basis is a cycle-consistency constraint, requiring that a cyclically translated
image should match the original. Cycle-consistent models can show remarkable performance~\cite{torbunov2023uvcgan}, but
there are concerns that the cycle-consistency condition may be too restrictive.

\aclgan~\cite{zhao2020unpaired} attempts to relax the cycle-consistency constraint and replace it with a weaker
adversarial one. Such relaxation allows the network to make larger changes to the source image, potentially achieving better translation quality. \council~\cite{nizan2020breaking} moves a step further and
completely discards cycle consistency. Instead, it trains an ensemble of generators, performing translation
in a single direction and allowing for a larger diversity of generated images.

\cut~\cite{park2020contrastive} takes an alternate route and uses a contrastive loss to maximize the information
between the source and the translated images. This approach removes the need to have multiple generators and allows
\cut to train faster. Using \cut as a basis, \ittr~\cite{zheng2022ittr} improves its performance by modifying the
generator architecture. In a similar fashion, \somesim~\cite{zheng2021spatially} designs a contrastive-based loss
function that guides the image translation without the need for multiple generators.

\paragraph{Diffusion-based Methods.} 
With the recent explosion of interest in DMs,
many have attempted to employ them
for unpaired I2I translation. For instance, \ilvr~\cite{choi2021ilvr} achieves an unpaired image translation
by modifying the standard Gaussian denoising process. It relies on a DM trained only on the
target domain but guides it toward the source image during denoising.

\sdedit~\cite{meng2021sdedit} introduces another viable approach for performing image translation. Instead
of modifying the diffusion process itself, it simply changes the starting point of diffusion. \sdedit uses a source
image perturbed by Gaussian noise as a seed image and runs the standard diffusion process on top of it.

Finally, \egsde~\cite{zhao2022egsde} makes an observation that both \ilvr and \sdedit are
trained on the target domain data. As such, they may perform a suboptimal translation. \egsde combines
\ilvr and \sdedit approaches and modifies both the starting point of the denoising process and the
denoising process itself. To overcome the limitation of the DM being trained only on the target
domain, it introduces a special energy function, pretrained on both domains. This energy function
guides the denoising process, allowing it to achieve \sota results on several benchmarks.

%% file: src/method.tex
\section{Method}

\thename revisits the classic \cyclegan~\cite{zhu2017unpaired} architecture.
\thename inherits several advancements from \uvcgan~\cite{torbunov2023uvcgan}, such as a hybrid U-Net-ViT (vision transformer) generator architecture, self-supervised generator pre-training,
and better training strategies.
This section describes several improvements that exceed UVCGAN, including the generator, discriminator and the training procedure.

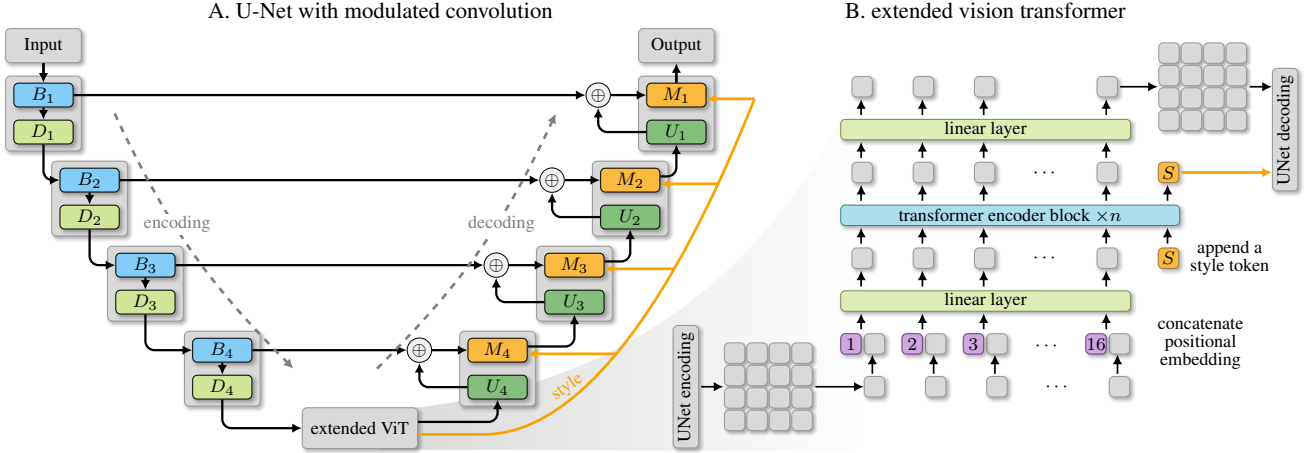
\begin{figure*}[t]
    \centering
    \resizebox{\textwidth}{!}{
        \tikzexternaldisable
        \tikzsetnextfilename{generator}%
        \begin{tikzpicture}
            \node[inner sep=1pt, fill=white] at (0, 0) {
                \begin{tikzpicture}
                    \node[inner sep=1pt] (unet) at (0, 0) {\input{fig/unet_modconv}};
                    \node[inner sep=1pt] (evit) at ([yshift=-3pt]$(unet.west)!1.3!(unet.east)$) {\input{fig/vit_extended}};
                    \node[inner sep=1pt, anchor=south] (unet_title) at ([yshift=2pt]unet.north) {A.~U-Net with modulated convolution};
                    \node[inner sep=1pt, anchor=south] (evit_title) at (unet_title.south -| evit) {B.~extended vision transformer};
                    \coordinate (unet_up) at ($(unet.north)!.905!(unet.south)$);
                    \coordinate (unet_dn) at ($(unet.north)!.995!(unet.south)$);
                    \coordinate (evit_up) at ([yshift=-20pt]$(evit.north west)!.3!(evit.north east)$);
                    \coordinate (evit_dn) at ($(evit.south west)!.3!(evit.south east)$);
                    \begin{pgfonlayer}{back}
                        \draw[name path=U, white] (unet_up) to[bend right=30] (evit_up);
                        \draw[name path=D, white] (unet_dn) -- (evit_dn);
                        \tikzfillbetween[of=U and D]{path fading=east, fill=gray!30}
                    \end{pgfonlayer}
                \end{tikzpicture}
            };
        \end{tikzpicture}
    }
    \caption{
        \textbf{\thename Generator.} 
        \thename's generator is a U-Net (Panel A) with an extended vision transformer bottleneck (eViT, Panel B). 
        The eViT outputs a style token for the modulated convolution blocks~\cite{karras2020analyzing} ($M_i$, $i=1,2,3,4$) in the decoding path of the U-Net.
        Refer to \cite{torbunov2023uvcgan} for details about the input layer, output layer, basic block $B_i$, downsampling block $D_i$ and upsampling block $U_i$ in the U-Net, and transformer encoder block in the eViT.
    }
    \label{fig:uvcgan2_generator}
\end{figure*}

\subsection{Review of the Original \uvcgan}
\label{subsec:cyclegan}

\uvcgan follows the \cyclegan framework~\cite{zhu2017unpaired,kim2019u}
that interlaces two generator-discriminator pairs for unpaired I2I translation.
Denote the two domains by $A$ and $B$. 
Generator $\gab$ translates images in $A$ to resemble those from domain $B$. 
Discriminator $\db$ distinguishes images in $B$ from those translated from $A$. 
The same goes for the other translation direction, $\gba$ and $\da$.
The discriminators are updated by backpropagating loss in distinguishing real and translated (fake) images (called \textit{GAN loss}):
\newcommand{\gl}{\ell_\text{gan}}
\begin{align}
    \mathcal{L}^{\text{disc}}_{A} &= \mathbb{E}_{B}\gl\paren{\da\paren{\fake{a}}, 0} + \mathbb{E}_{A}\gl\paren{\da(a), 1}, \label{eq:disc_loss_A}\\
    \mathcal{L}^{\text{disc}}_{B} &= \mathbb{E}_{A}\gl\paren{\db\paren{\fake{b}}, 0} + \mathbb{E}_{B}\gl\paren{\db(b), 1} \label{eq:disc_loss_B}
\end{align}
where $a$ is an image from $A$, $b$ is an image from $B$, $\fake{b}= \gab(a)$, $\fake{a} = \gba(b)$ (subscript $\text{f}$ means fake),
$0$ is the label for fake images, $1$ is the label for real images, 
and $\gl$ represents a classification loss function ($L_2$, cross-entropy, Wasserstein~\cite{arjovsky2017wasserstein}, etc.). 
The generators are updated by backpropagating loss from multiple sources: GAN loss for realistic translation, cycle-consistency loss, and optionally identity loss for within-domain translation. 
Using domain $A$ as an example, we have:
\begin{align}
    \mathcal{L}^{\text{gan}}_{A} &= \mathbb{E}_{A}\gl\paren{\db\paren{\fake{b}}, 1},\\
    \mathcal{L}^{\text{cyc}}_{A} &= \mathbb{E}_{A}\ell_\text{reg}\paren{\reco{a}, a}, \quad \mathcal{L}^{\text{idt}}_{A} = \mathbb{E}_{A}\ell_\text{reg}\paren{\iden{a}, a}
\end{align}
where $\reco{a} = \gba\circ\gab(a)$, $\iden{a} = \gba(a)$, and $\ell_\textrm{reg}$ is any pixel-wise loss function ($L_1$ or $L_2$, etc.). The generator loss is defined as a linear combination:
\newcommand{\nplus}{\!+\!}
\begin{equation}
    \mathcal{L}^{\text{gen}} \!=\! \paren{\mathcal{L}^{\text{gan}}_{A} \nplus \mathcal{L}^{\text{gan}}_{B}} \nplus \lambda^\text{c}\!\paren{\mathcal{L}^{\text{cyc}}_{A} \nplus \mathcal{L}^{\text{cyc}}_{B}} \nplus \lambda^\text{i}\!\paren{\mathcal{L}^{\text{idt}}_{A} \nplus \mathcal{L}^{\text{idt}}_{B}} \label{eq:gen_loss}
\end{equation}

\subsection{Source-driven Style Modulation}
\label{subsec:modulation}

\uvcgan uses a hybrid UNet-ViT generator network.
We attempt to increase the generator's performance by redesigning its architecture.
We extend the generator, enabling it to infer an appropriate target style for each input image.
Then, we modulate~\cite{karras2020analyzing} the generator's decoding branch by the style, significantly increasing its expressiveness.

Specifically, at the bottleneck of the generator, the image is encoded as a sequence of tokens to be fed to the Transformer network. 
We augment this sequence with an additional learnable style token $S$.
The state of the $S$ token at the output of the Transformer serves as a latent image style.
For each convolutional layer of the U-Net's decoding branch, we generate a specific
style vector $s_i$ from $S$ by trainable linear transformations.

The style modulation~\cite{karras2020analyzing} effectively scales weights $w_{i, j, x, y}$ of the convolutional operator
by the supplied style vector $s_i$, yielding modulated weights:
\begin{equation}
    w_{i, j, x, y}^\prime = s_{i} \cdot w_{i, j, x, y}
    \label{eq:weight_modulation}
\end{equation}
where $i$, $j$ refer to the input and output feature maps and $x$, $y$ enumerate the spatial dimensions.
To preserve the magnitude of the activations, the scaled weights $w_{i, j, x, y}^\prime$ need
to be demodulated. 
The demodulation operation further renormalizes the 
convolution weights as follows:
\begin{equation}
    w_{i, j, x, y}^{\prime\prime} =
        \frac{w_{i, j, x, y}^{\prime}}
            {\sqrt{\sum_{i, x, y} \paren{w_{i, j, x, y}^{\prime}}^2 + \epsilon}}
    \label{eq:weight_demodulation}
\end{equation}
where $\epsilon$ is a small number to prevent numerical instability.

Our approach is different from \styleganTwo, which generates the style vectors $s_i$ from a random prior. We use a learnable $S$ token of the transformer to infer the required target style from the source image itself. The way $S$ is processed is similar to the \texttt{[class]} token of the ViT~\cite{dosovitskiy2020image}. However the \texttt{[class]} token mainly is used for the classification task.

\subsection{Batch-Statistics-aware Discriminator}

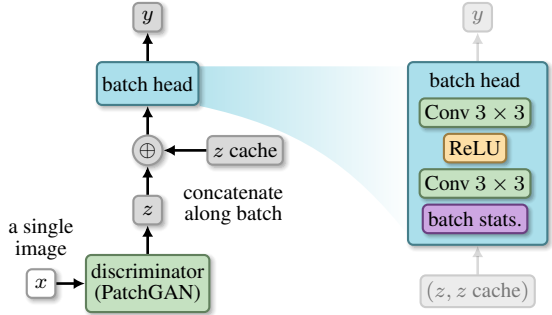
\begin{figure}[t]
    \centering
    \tikzexternaldisable
    \tikzsetnextfilename{batchhead}%
    \begin{tikzpicture}
        \node[inner sep=0, fill=white] at (0, 0) {\input{fig/disc_batchhead}};
    \end{tikzpicture}
    \caption{\textbf{\thename Discriminator with a batch head.} 
             For the batch statistics block, we can use either a standard batch normalization layer or batch standard deviation~\cite{karras2017progressive}.}
    \label{fig:batch_head}
\end{figure}

Batch statistics offers an effective way to combat mode collapse and
promote diversity for GAN models~\cite{salimans2016improved,karras2017progressive}.
We seek to improve the \uvcgan discriminator's performance by integrating a minibatch discrimination technique. However, employing traditional minibatch discrimination requires having a nontrivial batch size. With the recent push to
work on high-resolution images, increasing the batch size will strain GPU hardware
limitations. We workaround this problem by decoupling the batch size from the minibatch
discrimination technique and designing a GPU memory-efficient algorithm.

The main idea is to integrate a cache of past discriminator features to serve as a 
substitute for a large batch size.
Similar ideas, known as ``memory bank,'' have been explored in representation
learning models~\cite{he2020momentum,wu2018unsupervised}.
To our best knowledge, using a memory bank type of cache in a GAN is novel.

We design a composite discriminator made of a main \textit{body} and a \textit{batch head}.
The composite discriminator body can be any common discriminator, but, for the purposes of this work, we use \uvcgan's discriminator without the last layer.
The batch head is made of a layer that captures batch statistics, followed by two convolutional layers
(\autoref{fig:batch_head}).
This modular discriminator architecture allows for easily swapping different
discriminator bodies while still preserving the minibatch discrimination power.
During training, the batch features are stored in four separate
caches: real images from domain~\domA, real images from domain~\domB, and fake
images from both domains. All the caches have a fixed size and follow the
first-in-first-out (FIFO) update policy. 

The discriminator's batch head receives a concatenation (along the batch
dimension) of the discriminator body output for the current minibatch along
with a history of the past outputs from a cache (\autoref{fig:batch_head}). 
We experiment with two types of batch statistics layers: batch standard deviation
(BSD)~\cite{karras2017progressive} and a simple batch
normalization (BN).

\subsection{Pixel-wise Consistency Loss}
\label{subsec:consistLoss}
To improve the consistency of the generated and source images, we experiment with the addition of
an extra term $\mathcal{L}_\text{consist}$ to denote generator loss~\eqref{eq:gen_loss}. This term captures
the $L_1$ difference between the downsized versions of the source and translated images.
For example, for images of domain $A$
\begin{equation}
    \mathcal{L}_{\text{consist}, A} = \mathbb{E}_{A}\ell_1\paren{F(\gab(a)), F(a)}  
\end{equation}
where $F$ is a resizing operator down to $32 \times 32$ pixels (low-pass filter).
We add this term to the generator loss~\eqref{eq:gen_loss} with a magnitude $\lambda_\text{consist}$
for both domains.

\subsection{Modern Training Techniques}
\label{subsec:TrainingTech}
\uvcgan and \cyclegan use outdated GAN training techniques. 
Hence, we revamp the training procedure with a few modern additions.
First, we implement the exponential averaging of the generator weights~\cite{karras2017progressive}, 
making the generator less dependent on random fluctuations occurring in the GAN training.
Second, we use spectral normalization of the discriminator weights~\cite{miyato2018spectral},
enhancing the stability of the discriminator and training procedure as a whole.
Third, we experiment with using unequal learning rates for the generator
and discriminator~\cite{heusel2017gans},
which has been empirically shown to improve model performance.
Finally, we replace the generic gradient penalty (GP) with an improved zero-centered
GP version~\cite{thanh2019improving}, developed to promote the GAN's convergence.

%% file: fig/unet_modconv.tex
    \begin{tikzpicture}[font=\fontsize{8}{8}\selectfont]
        \tikzstyle{block} = [
            rectangle, 
            rounded corners=2, 
            blur shadow={
                shadow blur steps=5, 
                shadow xshift=1pt, 
                shadow yshift=-1pt
            },
            align=center,
            draw=black!40, 
        	fill=black!15,
        ]

        \def\xs{145pt}
        \def\ys{150pt}
        \def\lw{1pt}
        
        \begin{scope}[x=\xs, y=\ys]

            \tikzstyle{encdec} = [block, minimum width=.23 * \xs, minimum height=.22 * \ys]
            \tikzstyle{inout} = [block, minimum width=.23 * \xs, minimum height=.1 * \ys]
            \tikzstyle{inner} = [inner sep=1pt, block, minimum width=.18 * \xs, minimum height=.07 * \ys]
            \tikzstyle{basic} = [inner, draw=b1!.5!black, fill=b1!50]
            \tikzstyle{downsample} = [inner, draw=yg!.5!black, fill=yg!50]
            \tikzstyle{upsample} = [inner, draw=pg!.5!black, fill=pg!75]
            \tikzstyle{modconv} = [inner, draw=or!.5!black, fill=or!75]
            \tikzstyle{bottleneck} = [block, minimum width=.35 * \xs, minimum height=.12 * \ys]
            \tikzstyle{concat} = [circle, inner sep=.005 * \xs, draw=black, fill=black!5]
            \tikzstyle{flow} = [-{Latex[length=1.5mm,width=1.5mm]}, rounded corners, line width=\lw]

            \node[bottleneck] (b) at (0, 0) {extended ViT};
            \foreach \layerId/\level in {4/.7, 3/1.7, 2/2.7, 1/3.7} {
                \pgfmathsetmacro\height{\level / 4}
                \pgfmathparse{sqrt(\height)}
                \pgfmathsetmacro{\width}{\pgfmathresult}

                \node[encdec] (enc_\layerId) at (-\width, \height) {};
                \node[encdec] (dec_\layerId) at (\width, \height) {};
                \def\innershift{.018}
                \node[basic, anchor=south] (b_\layerId) at (-\width, \height + \innershift) {$B_{\layerId}$};
                \node[downsample, anchor=north] (d_\layerId) at (-\width, \height - \innershift) {$D_{\layerId}$};
                \node[upsample, anchor=north] (u_\layerId) at (\width, \height - \innershift) {$U_{\layerId}$};
                \node[modconv, anchor=south] (m_\layerId) at (\width, \height + \innershift) {$M_{\layerId}$};
                \node[concat] (c_\layerId) at ($(m_\layerId.east)!1.8!(m_\layerId.west)$) {$\oplus$};
                
                \draw[flow] (b_\layerId) -- (d_\layerId);
                \draw[flow] (b_\layerId) -- (c_\layerId);
                \draw[flow] (u_\layerId) -| (c_\layerId);
                \draw[flow] (c_\layerId) -- (m_\layerId);
            }
            
            \draw[flow] (d_4) |- (b);
            \draw[flow] ($(b.north east)!.35!(b.south east)$) -| (u_4);
            \foreach \fr/\to in {1/2, 2/3, 3/4} {
                \draw[flow] (d_\fr) |- (b_\to);
                \draw[flow] ($(m_\to.north east)!.35!(m_\to.south east)$) -| (u_\fr);
            }
            
            \def\modCol{or}
            \def\bendPointXShift{.2 * \xs}
            \def\lw{1.2pt}
            \draw[line width=\lw, draw=\modCol] ($(b.north east)!.65!(b.south east)$) --++ (\bendPointXShift, 0);
            \node[inner sep=1pt, text=\modCol, anchor=south west, rotate=35] at ($(b.north west)!2.2!(b.north east)$) {style};
            \coordinate (l) at ($(m_1.north west)!.65!(m_1.south west)$);
            \coordinate (r) at ($(m_1.north east)!.65!(m_1.south east)$);
            
            \draw[draw=\modCol, name path=parabola, line width=\lw] ([xshift=\bendPointXShift]$(b.north east)!.65!(b.south east)$) parabola ($(l)!1.8!(r)$);
            \foreach \layerId in {1, 2, 3, 4} {
                \coordinate (p_\layerId) at ($(m_\layerId.north east)!.65!(m_\layerId.south east)$);
                \path[name path=phantom, overlay] (p_\layerId) --++ (.5 * \xs, 0);
                \draw[flow, draw=\modCol, name intersections={of=parabola and phantom, by=i}] (i) -- (p_\layerId); 
            }

            \node[inout] (input) at ($(b_1.south)!2.5!(b_1.north)$) {Input};
            \node[inout] (output) at ($(m_1.south)!2.5!(m_1.north)$) {Output};
            \draw[flow] (input) -- (b_1);
            \draw[flow] (m_1) -- (output);

            \draw[flow,dashed,black!50] ([xshift=.1 * \xs]enc_1.east) to[bend right=10] ([xshift=0.1 * \xs]enc_4.east);
            \node[inner sep=1pt,align=center,text=black!50, fill=white] at ([xshift=0.15 * \xs, yshift=-.075 * \ys]enc_2.east) {encoding};
            \draw[flow,dashed,black!50] ([xshift=-.25 * \xs, yshift=0]dec_4.west) to[bend right=10] ([xshift=-.25 * \xs]dec_1.west);
            \node[inner sep=1pt,align=center,text=black!50,fill=white] at ([xshift=-.275 * \xs, yshift=-.075 * \ys]dec_2.west) {decoding};
        \end{scope}
    \end{tikzpicture}

%% file: fig/vit_extended.tex
    
    \begin{tikzpicture}[font=\fontsize{8}{8}\selectfont]
        \tikzstyle{block} = [
            inner sep=0,
            rectangle, 
            rounded corners=2, 
            blur shadow={
                shadow blur steps=5, 
                shadow xshift=1pt, 
                shadow yshift=-1pt
            },
            align=center,
            draw=black!40, 
        	fill=black!15,
        ]

        \tikzset{
            shadowed/.style={
                preaction={
                    transform canvas={shift={(1pt, -1pt)}},
                    draw=gray,
                    opacity=.75,
                    line width=1pt
                }
            },
        }
        
        \def\xs{180pt}
        \def\ys{200pt}
        \def\lw{.75pt}
        \begin{scope}[x=\xs, y=\ys]
            \def\s{.05 * \xs}
            \tikzstyle{token} = [block, minimum size=\s]
            \tikzstyle{style} = [block, minimum size=\s, draw=or!50!black!75, fill=or!75]
            \tikzstyle{posemb} = [block, minimum size=\s, draw=pr!50!black!75, fill=pr!35]
            \tikzstyle{linear} = [block, draw=yg!50!black!75, fill=yg!35]
            \tikzstyle{transformer} = [block, draw=b2!50!black!75, fill=b2!35]
            \tikzstyle{flow} = [-{Latex[length=1.5mm,width=1.5mm]}, rounded corners, line width=\lw, shorten <=1pt, draw]
            
            \def\hexp{3}
            \def\vexp{1.05}

            \foreach \p/\c in {1/1, 2/2, 3/3, 16/5} {
                \node[token] (i_\c) at (\c * \s * \hexp, 0) {}; 
                \node[token] (ip_\c) at (\c * \s * \hexp, 2 * \vexp * \s) {};
                \path[flow] ([xshift=-.1 * \s]i_\c.north) -- ([xshift=-.1 * \s]ip_\c.south);
                \node[posemb] (p_\c) at (\c * \s * \hexp - 1.15 * \s, 2 * \vexp * \s) {$\p$};

                \node[inner sep=0] (l1it_\c) at (\c * \s * \hexp - .65 * \s, 2.5 * \vexp * \s) {};
                \node[inner sep=0] (l1ih_\c) at (\c * \s * \hexp - .65 * \s, 3.5 * \vexp * \s) {};
                \path[flow] (l1it_\c) -- (l1ih_\c);
                \node[inner sep=0] (l1ot_\c) at (\c * \s * \hexp - .65 * \s, 4.5 * \vexp * \s) {};
                \node[inner sep=0] (l1oh_\c) at (\c * \s * \hexp - .65 * \s, 5.5 * \vexp * \s) {};
                \path[flow] (l1ot_\c) -- (l1oh_\c);
                
                \node[token] (tin_\c) at (\c * \s * \hexp - .6 * \s, 6 * \vexp * \s) {};
                \node[token] (tout_\c) at (\c * \s * \hexp - .6 * \s, 10 * \vexp * \s) {};

                \node[inner sep=0] (tit_\c) at (\c * \s * \hexp - .65 * \s, 6.5 * \vexp * \s) {};
                \node[inner sep=0] (tih_\c) at (\c * \s * \hexp - .65 * \s, 7.5 * \vexp * \s) {};
                \path[flow] (tit_\c) -- (tih_\c);
                \node[inner sep=0] (tot_\c) at (\c * \s * \hexp - .65 * \s, 8.5 * \vexp * \s) {};
                \node[inner sep=0] (toh_\c) at (\c * \s * \hexp - .65 * \s, 9.5 * \vexp * \s) {};
                \path[flow] (tot_\c) -- (toh_\c);

                \node[inner sep=0] (l2it_\c) at (\c * \s * \hexp - .65 * \s, 10.5 * \vexp * \s) {};
                \node[inner sep=0] (l2ih_\c) at (\c * \s * \hexp - .65 * \s, 11.5 * \vexp * \s) {};
                \path[flow] (l2it_\c) -- (l2ih_\c);
                \node[inner sep=0] (l2ot_\c) at (\c * \s * \hexp - .65 * \s, 12.5 * \vexp * \s) {};
                \node[inner sep=0] (l2oh_\c) at (\c * \s * \hexp - .65 * \s, 13.5 * \vexp * \s) {};
                \path[flow] (l2ot_\c) -- (l2oh_\c);

                \node[token] (o_\c) at (\c * \s * \hexp - .6 * \s, 14 * \vexp * \s) {};
                
            }
            \node[inner sep=0, anchor=west, align=center] (plabel) at (6 * \s * \hexp - 1.15 * \s, 2 * \vexp * \s) {concatenate\\positional\\embedding};
            
            \node[inner sep=0] at (4 * \s * \hexp, 0) {$\cdots$};
            \node[inner sep=0] at (4 * \s * \hexp - .5 * \s, 2 * \s) {$\cdots$};
            \node[inner sep=0] at (4 * \s * \hexp - .6 * \s, 6 * \vexp * \s) {$\cdots$};
            \node[inner sep=0] at (4 * \s * \hexp - .6 * \s, 10 * \vexp * \s) {$\cdots$};

            \node[style] (sin) at (6 * \s * \hexp - .6 * \s, 6 * \vexp * \s) {$S$};
            \node[inner sep=0, align=center] (slabel) at (7 * \s * \hexp - .6 * \s, 6 * \vexp * \s) {append a\\style token};
            \node[style] (sout) at (6 * \s * \hexp - .6 * \s, 10 * \vexp * \s) {$S$};
            
            \node[inner sep=0] (tit) at (6 * \s * \hexp - .65 * \s, 6.5 * \vexp * \s) {};
            \node[inner sep=0] (tih) at (6 * \s * \hexp - .65 * \s, 7.5 * \vexp * \s) {};
            \path[flow] (tit) -- (tih);
            \node[inner sep=0] (tot) at (6 * \s * \hexp - .65 * \s, 8.5 * \vexp * \s) {};
            \node[inner sep=0] (toh) at (6 * \s * \hexp - .65 * \s, 9.5 * \vexp * \s) {};
            \path[flow] (tot) -- (toh);

            \tikzmath{coordinate \L;\L = (ip_5.north east) - (p_1.south west);}
            \coordinate (lm) at ($(l1it_1)!.5!(l1oh_1)$);
            \node[linear, minimum width=\Lx, minimum height=\s, anchor=west] (linear1) at (p_1.west |- lm) {linear layer};
            \coordinate (lm) at ($(l2it_1)!.5!(l2oh_1)$);
            \node[linear, minimum width=\Lx, minimum height=\s, anchor=west] (linear2) at (p_1.west |- lm) {linear layer};
            \tikzmath{coordinate \T;\T = (sin.north east) - (linear1.south west);}
            \coordinate (lm) at ($(tit_1)!.5!(toh_1)$);
            \node[transformer, minimum width=\Tx, minimum height=\s, anchor=west] (trans) at (p_1.west |- lm) {transformer encoder block $\times n$};

            \newcommand{\im}{
                \begin{tikzpicture}
                    \def\exp{1.1}
                    \foreach \r in {1, 2, 3, 4} {
                        \foreach \c in {1, 2, 3, 4} {
                            \node[token] at (\c * \s * \exp, -\r * \s * \exp) {};
                        }
                    }    
                \end{tikzpicture}
            }
            \node[inner sep=0, anchor=east] (enOut) at (0, 0) {\im};
            \node[block, rotate=90, minimum height=\s, minimum width=6 * \s, anchor=south, inner sep=2] (unet_en) at ([xshift=-1.2 * \s]enOut.west) {UNet encoding};
            \path[flow] (unet_en) -- (enOut);
            \node[inner sep=0, anchor=west] (deIn) at (sout.west |- o_5) {\im};
            \path[flow] (enOut) -- (i_1);
            \path[flow] (o_5) -- (deIn);
            \coordinate (de) at ([xshift=1.2 * \s]deIn.east);
            \path[flow] (deIn.east) -- (de);
            \path[flow, draw=or, line width=1.2pt] (sout) -- (de |- sout);
            \coordinate (u) at ($(deIn.east)!.5!(sout.east)$) {};
            \node[block, rotate=90, minimum height=\s, minimum width=6 * \s, anchor=north, inner sep=2] (unet_de) at (u -| de) {UNet decoding}; 
        \end{scope}
    \end{tikzpicture}

%% file: fig/disc_batchhead.tex
    \begin{tikzpicture}[font=\fontsize{8}{8}\selectfont]
        \tikzstyle{block} = [
            thick,
            inner sep=2pt,
            rectangle, 
            rounded corners=2, 
            blur shadow={
                shadow blur steps=5, 
                shadow xshift=1pt, 
                shadow yshift=-1pt
            },
            align=center,
            draw=black!40, 
        	fill=black!15,
        ]

        \def\xs{180pt}
        \def\ys{200pt}
        \def\lw{1pt}
        
        \begin{scope}[x=\xs, y=\ys]
            \def\s{.06 * \xs}
            \tikzstyle{data} = [block, minimum size=\s]
            \tikzstyle{queue} = [block, minimum width=.1 * \xs, minimum height=\s]
            \tikzstyle{disc} = [block, minimum width=.1 * \xs, minimum height=2 * \s, draw=pg!50!black!75, fill=pg!35]
            \tikzstyle{batchHead} = [block, minimum width=.12 * \xs, minimum height=1.5 * \s, draw=b2!50!black!75, fill=b2!35]
            \tikzstyle{batchHeadDetail} = [block, draw=b2!50!black!75, fill=b2!35]
            \tikzstyle{concat} = [block, circle, inner sep=.005 * \xs]
            \tikzstyle{conv} = [block, minimum width=.1 * \xs, minimum height=\s, draw=pg!50!black!75, fill=pg!35]
            \tikzstyle{activ} = [block, minimum width=.1 * \xs, minimum height=\s, draw=or!50!black!75, fill=or!35]
            \tikzstyle{batchStats} = [block, minimum width=.1 * \xs, minimum height=\s, draw=pr!50!black!75, fill=pr!35]
            \tikzstyle{flow} = [-{Latex[length=1.5mm,width=1.5mm]}, rounded corners, line width=\lw]

            \node[disc,align=center] (disc) at (0, 0) {discriminator\\(PatchGAN)};
            \node[data, anchor=east, fill=white] (x) at ([xshift=-\s]disc.west) {$x$};
            \node[inner sep=0, anchor=south, align=center] (xlabel) at ([yshift=.3 * \s]x.north) {a single\\image};
            \node[data, anchor=south] (z) at ([yshift=\s]disc.north) {$z$};
            \node[concat, anchor=south] (p) at ([yshift=\s]z.north) {$\oplus$};
            \node[queue, anchor=west] (zq) at ([xshift=1.5 * \s]p.east) {$z$ cache};
            \node[batchHead, anchor=south] (bh) at ([yshift=\s]p.north) {batch head};
            \node[inner sep=0, align=center, anchor=south east] (plabel) at (zq.east |- z.south) {concatenate\\along batch};
            \node[data, anchor=south] (y) at ([yshift=\s]bh.north) {$y$};
            
            \draw[flow] (x) -- (disc);
            \draw[flow] (disc) -- (z);
            \draw[flow] (z) -- (p);
            \draw[flow] (zq) -- (p);
            \draw[flow] (p) -- (bh);
            \draw[flow] (bh) -- (y);

            \node[batchHeadDetail, anchor=north west, inner sep=4pt] (bhd) at ($(bh.north west)!3!(bh.north east)$) {
                \begin{tikzpicture}
                    \def\yshift{.01 * \xs}
                    \node[batchStats] (bs) at (0, 0) {batch stats.};
                    \node[conv, anchor=south] (conv1) at ([yshift=\yshift]bs.north) {Conv $3\times 3$};
                    \node[activ, anchor=south] (activ) at ([yshift=\yshift]conv1.north) {ReLU};
                    \node[conv, anchor=south] (conv2) at ([yshift=\yshift]activ.north) {Conv $3\times 3$};
                    \node[inner sep=0, anchor=south] (label) at ([yshift=2 * \yshift]conv2.north) {batch head};
                \end{tikzpicture}
            };
            \def\op{.5}
            \node[queue, anchor=north, opacity=\op, blur shadow={shadow opacity=.2}] (bh_input) at ([yshift=-\s]bhd.south) {$(z, z\textrm{ cache})$};
            \node[data, anchor=south, opacity=\op, blur shadow={shadow opacity=.2}] (bh_output) at ([yshift=\s]bhd.north) {$y$};
            \draw[flow, opacity=.3 * \op] (bh_input) -- (bhd);
            \draw[flow, opacity=.3 * \op] (bhd) -- (bh_output);

            \begin{pgfonlayer}{back}
                \draw[name path=U, white] (bh.south east) to[bend left=20] ([xshift=2pt, yshift=2pt]bhd.south west);
                \draw[name path=D, white] ([yshift=-2pt]bh.north east) -- ([yshift=-2pt]bhd.north west);
                \tikzfillbetween[of=U and D]{black!30, path fading=east, fill=b2!30}
            \end{pgfonlayer}
        \end{scope}
    \end{tikzpicture}

%% file: src/experiments.tex
\section{Experiments}

\subsection{Datasets}

We study the performance of \thename on two groups of datasets: the formerly widely used \celeba~\cite{liu2015deep}
and \anime~\cite{kim2019u} datasets and the modern, high-quality \celebahq~\cite{karras2017progressive} and
\afhq~\cite{choi2020stargan} datasets. 
More details about these datasets can be found in \autoref{sec:DatasetDetail}.

\paragraph{\celeba and \anime.} 
The \celeba and \anime datasets have been commonly used to benchmark GAN-based unpaired I2I translation
algorithms~\cite{torbunov2023uvcgan,zhu2017unpaired,nizan2020breaking,zhao2020unpaired,kim2019u}.
We study \thename performance on three tasks related to the \celeba and \anime datasets: \gender translation
on the \celeba dataset, Glasses Removal on the \celeba dataset, and Selfie-to-Anime translation on the \anime dataset.
Because the \cyclegan setup learns translations in both directions simultaneously,
we also get benchmarks in the opposite directions (Female-to-Male, \glasses Addition,
and Anime-to-Selfie translations).

\paragraph{\celebahq and \afhq.} 
To compare the performance of the \thename against more recent unpaired I2I translation
algorithms, such as EGSDE~\cite{zhao2022egsde}, we also consider the \celebahq and \afhq datasets. 
We investigate \malefemale translation on \celebahq
and three translations, \catdog, \wilddog, and \wildcat, on \afhq.

\paragraph{Preprocessing.} 
For a fair comparison with EGSDE~\cite{zhao2022egsde}, we downsize the \celebahq and \afhq images
to $256 \times 256$ pixels. To avoid Fr\'{e}chet inception distance (FID) evaluation inconsistencies associated with variations
in the interpolation procedures between different frameworks~\cite{parmar2022aliased}, we use the 
Pillow~\cite{pillow} image manipulation library to perform the image resizing with Lanczos
interpolation.

\subsection{Training}

When training our modified \uvcgan implementation, we seek to closely follow the original
two-step procedure~\cite{torbunov2023uvcgan}. The first involves pre-training the generator
in a self-supervised way on a task of image inpainting, while the second step is the actual training of
the unpaired I2I translation networks, starting from the pre-trained generators.

\paragraph{Generator Pre-training.} 
For each dataset, the generators are pre-trained on image inpainting tasks. This task is set up
in a fashion similar to the Bidirectional Encoder Representations from Transformers (BERT) pre-training~\cite{devlin2018bert,torbunov2023uvcgan}. For the inpainting task,
input images of size $256 \times 256$ pixels are tiled into a grid of patches at $32 \times 32$ pixels.
Then, each patch is masked with a probability of $40\%$. The masking is performed by zeroing out
pixel values. The generator is tasked to recover the original unmasked image from a masked one. More details about this pre-training are available in \autoref{sec:TrainingDetails}.

\paragraph{Translation Training.} 
The unpaired image translation training is performed for one million iterations using the Adam optimizer. 
Depending on the dataset, we employ various data augmentations. 
For the preprocessed datasets, such as \celebahq and \afhq, only a random horizontal flip is applied.
For the \anime and \celeba datasets, we use three
augmentations: resize, followed by a random crop of size $256 \times 256$ then, finally, a random
horizontal flip. Resizing for the \anime dataset is done from $256 \times 256$ up to
$286 \times 286$. For the \celeba dataset, the resizing is done from $178 \times 218$ to $256 \times 313$.

\paragraph{Hyperparameter Tuning.} 
For each translation, we perform a quick hyperparameter search, exploring the space of the
cycle-consistency magnitude $\lambda_\text{cycle}$, magnitude of the zero-centered GP $\lambda_\text{GP}$, magnitude of the consistency loss $\lambda_\text{consist}$, 
learning rates of the generator and discriminator, and the choice of the batch head (BN versus BSD).
We also explore turning the learning rate scheduler on and off. 
Refer to \autoref{sec:TrainingDetails} 
for more details.

%% file: src/results.tex
\section{Results}

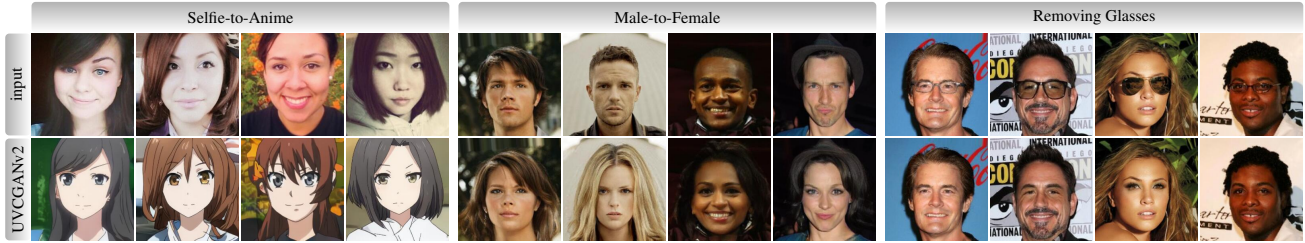
\begin{figure*}[t]
    \centering
    \tikzexternaldisable
    \tikzsetnextfilename{grid_LQ}
    \input{fig/grid}
    \resizebox{\textwidth}{!}{
    \begin{tikzpicture}
        
        \def\width{.1\textwidth}
        \def\expand{1.03}
        \def\folder{fig/grid_LQ}
        
        \node[inner sep=0, fill=white] at (0, 0) {
            \begin{tikzpicture}
                \node[inner sep=0] (G1) at (0, 0) {\gridWithHeader[1]{selfie2anime}{0, 1, 3, 6}{input/input, UVCGAN2/\thename}{\selfieanime}{4}};
            	\node[inner sep=0, anchor=west] (G2) at ([xshift=.03 * \width]G1.east) {\gridWithHeader{male2female}{4, 1, 3, 9}{input/input, UVCGAN2/\thename}{\malefemale}{4}};
            	\node[inner sep=0, anchor=west] (G3) at ([xshift=.03 * \width]G2.east) {\gridWithHeader{rmvGlasses}{4, 7, 8, 2}{input/input, UVCGAN2/\thename}{\rmvGlasses}{4}}; 
            \end{tikzpicture}
        };
    \end{tikzpicture}
    }
    \caption{\textbf{Sample translations for \celeba and \anime. } Translations produced by \thename for three tasks: \selfieanime, \malefemale, and \rmvGlasses. The full grid with all benchmarking results 
    can be found in \autoref{sec:additionalTransSamples}.}
    \label{fig:grid_LQ}
\end{figure*}

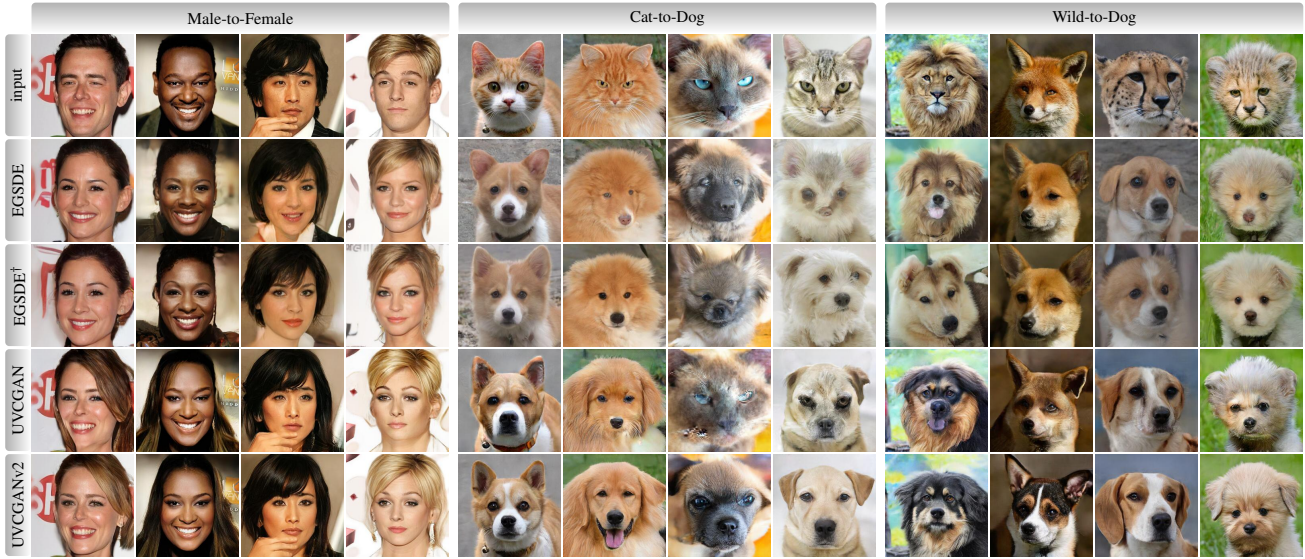
\begin{figure*}[t]
    \centering
    \tikzexternaldisable
    \tikzsetnextfilename{grid_HQ}
    \input{fig/grid}
    \resizebox{\textwidth}{!}{
    \begin{tikzpicture}
    
        \def\width{.1\textwidth}
        \def\expand{1.03}
        \def\folder{fig/grid_HQ}
    
        \node[inner sep=0, fill=white] at (0, 0) {
            \begin{tikzpicture}
                \node[inner sep=0] (G1) at (0, 0) {\gridWithHeader[1]{male2female}{1, 3, 5, 8}{input/input, EGSDE1/\egsde, EGSDE2/\egsdeDG, UVCGAN1/\uvcgan, UVCGAN2wo/\thename}{\malefemale}{4}};
            	\node[inner sep=0, anchor=west] (G2) at ([xshift=.03 * \width]G1.east) {\gridWithHeader{cat2dog}{9, 10, 11, 12}{input/input, EGSDE1/\egsde, EGSDE2/\egsdeDG, UVCGAN1/\uvcgan, UVCGAN2wo/\thename}{\catdog}{4}};
            	\node[inner sep=0, anchor=west] (G3) at ([xshift=.03 * \width]G2.east) {\gridWithHeader{wild2dog}{1, 3, 4, 5}{input/input, EGSDE1/\egsde, EGSDE2/\egsdeDG, UVCGAN1/\uvcgan, UVCGAN2wo/\thename}{\wilddog}{4}}; 
            \end{tikzpicture}
        };
    \end{tikzpicture}
    }
    \caption{\textbf{Sample translations for \celebahq and \afhq. } Translations for three tasks: \malefemale, \catdog, and \wilddog. More translations for these three tasks and those for \wildcat can be found in \autoref{sec:additionalTransSamples}.}
    \label{fig:grid_HQ}
\end{figure*}

\subsection{Metrics of Realism and Faithfulness}

There are two dimensions along which the unpaired I2I style transfer models can be evaluated:
\textit{Faithfulness} and \textit{Realism}. Faithfulness captures the degree of similarity between the
source and its translated image at an individual level. 
Realism attempts to estimate the overlap of the distributions
of the translated images and the ones in the target.

In terms of realism, image translation quality is commonly judged according to the
FID~\cite{heusel2017gans} and kernel inception distance
(KID)~\cite{binkowski2018demystifying} metrics. 
Both metrics measure the
distance between the distributions of the latent Inception-v3~\cite{szegedy2016rethinking}
features extracted from samples of the translated and target images. Smaller FID and KID values
indicate more realistic images.

Early GAN-based works (e.g.,~\cite{zhu2017unpaired,nizan2020breaking,zhao2020unpaired,kim2019u}) do not
explicitly evaluate the faithfulness of the translation. 
To the best of our knowledge, there is no widely accepted faithfulness metric available. 
Some works~\cite{zhao2022egsde} try to employ simple pixel-wise $L_2$, peak-signal-to-noise ratio (PSNR), or structural similarity index measure (SSIM)~\cite{wang2004image} scores to capture the agreement between the source and translation. 
Yet, it is unclear how well these pixel-wise metrics relate to the perceived image faithfulness,
and we explore more advanced alternatives in \autoref{sec:better_faithfulness}.

\subsection{Evaluation Protocol}

Evaluation protocols differ drastically between different papers (see \autoref{sec:RemarksMetricEvalConsistency}). 
This makes the direct comparison of the translation quality metrics extremely challenging.
For the fairness of comparisons with older works, we follow different evaluation protocols,
depending on the dataset.

\paragraph{\celeba and \anime.}
When evaluating the quality of translation on the \celeba \malefemale, \celeba \glasses Removal, and \anime datasets,
we use the evaluation protocol of \uvcgan~\cite{torbunov2023uvcgan}, which uniformized FID/KID evaluation across multiple datasets and models, allowing for a simple FID/KID comparison.
For the actual FID/KID evaluation, we rely on \texttt{torch-fidelity}~\cite{obukhov2020torchfidelity}, which provides
a validated implementation of these metrics.

The actual evaluation protocol for \celeba and \anime relies only on test splits to perform the FID/KID
evaluation. For the \celeba dataset, we use KID subset size of $1000$. 
For the \anime dataset, we use the KID subset size of $50$. 
We use unprocessed images of size $256 \times 256$ when evaluating on the \anime dataset. 
For the \celeba dataset, we apply a simple pre-processing to both domains: resizing
the smaller side to $256$ pixels then taking a center crop of size $256 \times 256$.

\paragraph{\celebahq and \afhq.} 
\egsde~\cite{zhao2022egsde} has evaluated multiple models on the \celebahq and \afhq datasets under
similar conditions. To compare our results to \egsde, we replicate its evaluation protocol for \celebahq 
and \afhq.
For the \afhq dataset, we evaluate FID and KID scores between the translated images of size $256 \times 256$ and
the target images of size $512 \times 512$ from the validation split.
For the \celebahq dataset, we evaluate the FID/KID scores between the translated images of size $256 \times 256$
and the downsized target images of size $256 \times 256$ from the train split. We perform the same channel
standardization as \egsde with $\mu = (0.485, 0.456, 0.406)$ and $\sigma = (0.229, 0.224, 0.225)$.
To ensure full consistency, we use the reference evaluation code provided by \egsde~\cite{egsderef}. 
\autoref{sec:consistent_eval} 
provides results of an alternative evaluation protocol that is uniform across all the datasets.

\subsection{Quantitative Results}

\paragraph{\celeba and \anime.}
\autoref{tab:results_lq} shows a comparison of the \thename (trained without piwel-wise consistency loss) performance against
\aclgan~\cite{zhao2020unpaired}, \council~\cite{nizan2020breaking},
\cyclegan~\cite{zhu2017unpaired}, \ugatit~\cite{kim2019u}, and
\uvcgan~\cite{torbunov2023uvcgan}.
The competitor models' performance is obtained from the \uvcgan paper~\cite{torbunov2023uvcgan}.
According to \autoref{tab:results_lq}, \thename outperforms all competitor models in all translation directions, except \animeselfie. The degree of improvement ranges from about $5\%$
in terms of FID on the \selfieanime translation to around $51\%$ on the \malefemale
translation. Likewise, there is a significant improvement in the KID scores from about
$13\%$ on \animeselfie to $79\%$ on \malefemale. This improvement demonstrates
the effectiveness of modern additions to the traditional \cyclegan architecture. \autoref{fig:grid_LQ} provides some translation samples, while more can be found in \autoref{sec:additionalTransSamples}. 

\input{table/results_lq}

\paragraph{\celebahq and \afhq.}
\autoref{tab:results_hq} compares the results of the \thename evaluation against \cut~\cite{park2020contrastive}, \ilvr~\cite{choi2021ilvr}, \sdedit~\cite{meng2021sdedit},
two versions of the \egsde~\cite{zhao2022egsde},
and the original \uvcgan~\cite{torbunov2023uvcgan} model.
In particular, this table compares two versions of the \thename: \thename and \mbox{\thename-C}. \mbox{\thename-C} is a version trained with a pixel-wise consistency loss and $\lambda_\text{consist} = 0.2$.
Here, we train \uvcgan models from scratch
(details are provided in \autoref{sec:app_train_uvcgan})
and extract the performance metrics of the remaining competitor models from \egsde~\cite{zhao2022egsde}.

\input{table/results_hq_v2}
\autoref{tab:results_hq} shows that \thename achieves the best FID scores across all translation tasks.
It outperforms EGSDE and UVCGAN by at least $10\%$ on both AFHQ tasks.
On the Male-to-Female task, it surpasses EGSDE by a significant amount, $43\%$.
Intriguingly, the original UVCGAN's performance on this task seems on par with UVCGANv2.
Yet, UVCGANv2 shows an advantage if we adapt the UVCGAN's evaluation protocol with $19.3\%$ improvement (FID 24.2 vs 30.0, c.f. \autoref{sec:app_train_uvcgan_consist}). 

The addition of the consistency loss allows the \thename-C
model to improve its pixel-wise PSNR and SSIM metrics---but at the expense of the FID score on \afhq translation.
\thename and \thename-C achieve competitive SSIM scores but lose in terms of the PSNR ratio
to the other models. However, as previously noted~\cite{zhang2018unreasonable}, pixel-wise measures PSNR and SSIM are not good metrics to judge perceptual image
faithfulness.
Overall, the gains in SSIM and PNSR scores provided
by the consistency loss to \thename-C do not seem to outweigh the associated FID losses.

Finally, it may be instructive to examine the diversity of the translated images.
We compare the diversity of the generated images according to a pairwise LPIPS distance~\cite{zhang2018unreasonable}, following an approach similar to~\cite{liu2021divco}.
To calculate the LPIPS distance, we use a VGG-based implementation (v0.0) in a consecutive pair mode.
\autoref{tab:results_hq} shows that \thename produces a much greater diversity of generated images compared to the \egsde variants,
yet it is slightly less diverse compared to \uvcgan.
It also indicates that the presence of the pixel-wise consistency loss
leads to a small but repeatable increase in image diversity.

\subsection{Ablation Study}

\input{table/app_ablation}

\autoref{tab:app_ablation} summarizes \thename ablation results on the \malefemale
translation of \celeba. To produce this table, we start with the final \thename configuration and make one of following modifications separately:
(a) disable style modulation in the generator;
(b) disable batch head of the discriminator;
(c) revert to \uvcgan training setup: linear scheduler, full GP, and disable generator averaging.

According to \autoref{tab:app_ablation}, the generator modifications (a) account for the majority of the performance improvement. Removing these modifications degrades the FID score of
the \malefemale translation from $4.7$ to $8.1$.
The discriminator modifications (b) provide a significant but relatively smaller improvement in the I2I performance. The new training setup (c) produces an improvement somewhere in between the generator and discriminator modifications.

\input{src/results_faithfulness}

%% file: fig/grid.tex
	\tikzstyle{header} = [
		inner sep=0, 
		anchor=south,
		rounded corners=2,
		fill=black!30,
		path fading=south,
        font=\fontsize{8}{8}\selectfont,
	]
	\tikzstyle{index} =[
		rotate=90,
		inner sep=0, 
		anchor=south,
		rounded corners=2,
		fill=black!30,
		path fading=east,
		font=\fontsize{8}{8}\selectfont,
	]

    \newcommand{\gridcp}[4][0]{
		\begin{tikzpicture}
			\foreach \name/\label [count=\row] in {#4} {
				\foreach \idx [count=\col] in {#3} {
					\def\fname{\folder/#2_\name_\idx.png.jpg}
                    \coordinate (C) at (\width * \expand * \col, -\width * \expand * \row);
					\node[inner sep=0, anchor=north] (\name_\col) at (C) {\includegraphics[width=\width]{\fname}};
                    \ifthenelse{\row=1}{
                        \pgfmathtruncatemacro\exampleNo{\col + #1}
                        \node[header, minimum width=\width, minimum height=.25 * \width, anchor=south] at (input_\col.north) {Example $\exampleNo$};}{}
				}
				\node[index, minimum width=\width, minimum height=.25 * \width] (index_\name) at (\name_1.west) {\label};
			}
		\end{tikzpicture}
	}
 
    \newcommand{\gridWithHeader}[6][0]{
        
        \begin{tikzpicture}
            \foreach \name/\label [count=\row] in {#4} {
                \foreach \idx [count=\col] in {#3} {
                    \def\fname{\folder/#2_\name_\idx.png.jpg}
                    \coordinate (Coor) at (\width * \expand * \col, -\width * \expand * \row);
                    \node[inner sep=0, anchor=north] (\name_\col) at (Coor) {\includegraphics[width=\width]{\fname}};
                }
                \ifthenelse{#1=1}{\node[index, minimum width=\width, minimum height=.25 * \width] (index_\name) at (\name_1.west) {\label}}{};
            }
            \tikzmath{ coordinate \M; \M = (input_#6.north east) - (input_1.south west); }
            \node[header, minimum width=\Mx, minimum height=.3 * \My, anchor=south west] at (input_1.north west) {#5};
        \end{tikzpicture}
	}
    \newcommand{\gridlm}[3][0]{
		\begin{tikzpicture}
            \pgfplotstableread[col sep=comma]{\folder/facelm_candidates.csv}\table
			\foreach \name/\label [count=\row] in {#3} {
				\foreach \idx [count=\col] in {#2} {
					\def\fname{\folder/\name_\idx.jpg}
                    \coordinate (C) at (2 * \width * \expand * \col - \width * \expand, -\width * \expand * \row);
					\node[inner sep=0, anchor=north] (\name_\col) at (C) {\includegraphics[width=\width]{\fname}};

                    \def\fnamelm{\folder/\name_\idx_lm.jpg}
                    \coordinate (C) at (2 * \width * \expand * \col, -\width * \expand * \row);
					\node[inner sep=0, anchor=north] (\name_\col_lm) at (C) {\includegraphics[width=\width]{\fnamelm}};
                    
                    \ifthenelse{\row=1}{
                        \pgfmathtruncatemacro\exampleNo{\col + #1}
                        \tikzmath{coordinate \C;\C = (input_\col_lm.north east) - (input_\col.south west);}
                        \node[header, minimum width=\Cx, minimum height=.25 * \width, anchor=south west] at (input_\col.north west) {Example $\exampleNo$};
                    }{
                        \pgfplotstablegetelem{\idx}{\name}\of\table
                        \pgfmathsetmacro\ll{\pgfplotsretval}
                        \node[fill=white, fill opacity=.7, text opacity=1, inner sep=2pt, anchor=north east] at (\name_\col_lm.north east) {\scriptsize$\ll$};
                    }
				}
				\node[index, minimum width=\width, minimum height=.25 * \width] (index_\name) at (\name_1.west) {\label};
			}
		\end{tikzpicture}
	}

%% file: table/results_lq.tex
\begin{table}[!ht]
    \caption{\textbf{FID and KID scores.} Lower is better.}
    \vskip-3mm
    \label{tab:results_lq}
    \resizebox{\linewidth}{!}{
    \centering
        \begin{tabular}{l|rr|rr}
            \toprule
            {} & \multicolumn{2}{r|}{Selfie-to-Anime} & \multicolumn{2}{r}{Anime-to-Selfie} \\
            {} & FID & KID ($\times100$) & FID & KID ($\times100$) \\ 
            \midrule
            \aclgan   & $99.3$ & $3.22\pm0.26$ & $128.6$ & $3.49\pm0.33$ \\
            \council  & $91.9$ & $2.74\pm0.26$ & $126.0$ & $2.57\pm0.32$ \\
            \cyclegan & $92.1$ & $2.72\pm0.29$ & $127.5$ & $2.52\pm0.34$ \\
            \ugatit   & $95.8$ & $2.74\pm0.31$ & $\mathbf{108.8}$ & \underline{$1.48\pm0.34$} \\
            \uvcgan   & \underline{$79.0$} & \underline{$1.35 \pm 0.20$}
                      & $122.8$ & $2.33\pm0.38$ \\
            \midrule
            \thename  & $\mathbf{75.8}$ & $\mathbf{1.18 \pm 0.28}$
                      & $\underline{113.8}$ & $\mathbf{1.26 \pm 0.23}$ \\
            \midrule
            \midrule
            {} & \multicolumn{2}{r|}{Male-to-Female} & \multicolumn{2}{r}{Female-to-Male} \\
            {} & FID & KID ($\times100$) & FID & KID ($\times100$) \\
            \midrule
            \aclgan   & \underline{$9.4$} & \underline{$0.58\pm0.06$} & $19.1$ & $1.38\pm0.09$ \\
            \council  & $10.4$ & $0.74\pm0.08$ & $24.1$ & $1.79\pm0.10$ \\
            \cyclegan & $15.2$ & $1.29\pm0.11$ & $22.2$ & $1.74\pm0.11$ \\
            \ugatit   & $24.1$ & $2.20\pm0.12$ & $15.5$ & $0.94\pm0.07$ \\
            \uvcgan   & $9.6$ & $0.68\pm0.07$ & \underline{$13.9$} & \underline{$0.91\pm0.08$} \\
            \midrule
            \thename  & $\mathbf{4.7}$ & $\mathbf{0.14 \pm 0.02}$ & $\mathbf{7.6}$ & $\mathbf{0.24 \pm 0.02}$ \\
            \midrule
            \midrule
            {} & \multicolumn{2}{r|}{Remove Glasses} & \multicolumn{2}{r}{Add Glasses} \\
            {} & FID & KID ($\times100$) & FID & KID ($\times100$) \\
            \midrule
            \aclgan   & $16.7$ & $0.70\pm0.06$ & $20.1$ & $1.35\pm0.14$ \\
            \council  & $37.2$ & $3.67\pm0.22$ &    $19.5$ & $1.33\pm0.13$ \\
            \cyclegan & $24.2$ & $1.87\pm0.17$ & $19.8$ & $1.36\pm0.12$ \\
            \ugatit   & $23.3$ & $1.69\pm0.14$ & $19.0$ & $1.08\pm0.10$ \\
            \uvcgan   & \underline{$14.4$} & \underline{$0.68\pm0.10$}
                      & \underline{$13.6$} & \underline{$0.60\pm0.08$} \\
            \midrule
            \thename  & $\mathbf{10.6}$ & $\mathbf{0.27 \pm 0.06}$ & $\mathbf{11.3}$ & $\mathbf{0.34 \pm 0.07}$ \\
            \midrule
            \bottomrule
        \end{tabular}
    }
    \vskip-3mm
\end{table}

%% file: table/results_hq_v2.tex
\begin{table}[t]
    \caption{\textbf{FID, PSNR, SSIM, and LPIPS scores.}}
    \vskip-3mm
    \label{tab:results_hq}
    \centering
    \resizebox{.85\linewidth}{!}{
    \begin{tabular}{l|rrrr}
        \toprule
        {} & \multicolumn{4}{c}{Male-to-Female (CelebA-HQ)}  \\
        {} & FID$\downarrow$ & PSNR$\uparrow$ & SSIM$\uparrow$ & LPIPS$\uparrow$ \\ 
        \midrule
        \cut             & $46.61$ & $19.87$ & \underline{$0.74$} & - \\
        \ilvr            & $46.12$ & $18.59$ & $0.510$ & -\\
        \sdedit          & $49.43$ & $20.03$ & $0.572$ & - \\
        \egsde  & $41.93$ & \underline{$20.35$} & $0.574$ & $0.162$ \\
        \egsdeDG & $30.61$ & $18.32$ & $0.510$ & $0.159$\\
        \uvcgan & \underline{$17.63$} & $19.30$ & $\mathbf{0.753}$ & $\mathbf{0.194}$ \\
        \midrule
        \thename  & $17.65$ & $19.44$ & $0.681$ & $0.188$ \\
        \thename-C
            & $\mathbf{17.34}$ & $\mathbf{21.18}$
            & \underline{$0.738$} & \underline{$0.190$} \\
        \midrule
        \midrule
        {} & \multicolumn{4}{c}{Cat-to-Dog} \\
        {} & FID$\downarrow$ & PSNR$\uparrow$ & SSIM$\uparrow$ & LPIPS$\uparrow$ \\
        \midrule
        \cut             & $76.21$ & $17.48$ & $0.601$ & -\\
        \ilvr            & $74.37$ & $17.77$ & $0.363$ & -\\
        \sdedit          & $74.17$ & \underline{$19.19$} & $0.423$ & - \\
        \egsde  & $65.82$ & $\mathbf{19.31}$ & $0.415$ & $0.199$ \\
        \egsdeDG & \underline{$51.04$}  & $17.17$ & $0.361$ & $0.196$ \\
        \uvcgan & $69.33$ & $18.36$ & $\mathbf{0.683}$ & $\mathbf{0.227}$ \\
        \midrule
        \thename & $\mathbf{44.76}$ & $15.55$ & $0.562$ & \underline{$0.221$} \\
        \thename-C & $52.48$ & $18.30$ & \underline{$0.638$} & \underline{$0.221$} \\
        \midrule
        \midrule
        {} & \multicolumn{4}{c}{Wild-to-Dog} \\
        {} & FID$\downarrow$ & PSNR$\uparrow$ & SSIM$\uparrow$ & LPIPS$\uparrow$ \\
        \midrule
        \cut             & $92.94$ & $17.2$ & $0.592$ & - \\
        \ilvr            & $75.33$ & $16.85$ & $0.287$ & -\\
        \sdedit          & $68.51$ & $17.98$ & $0.343$ & -\\
        \egsde   & $59.75$ & \underline{$18.14$} & $0.343$ & $0.193$\\
        \egsdeDG & \underline{$50.43$} & $16.40$ & $0.300$ & $0.190$ \\
        \uvcgan & $78.44$ & $17.67$ & $\mathbf{0.675}$ & $\mathbf{0.224}$ \\
        \midrule
        \thename & $\mathbf{45.56}$ & $15.59$ & $0.551$ & $0.217$ \\
        \thename-C & $55.61$ & $\mathbf{18.65}$ & \underline{$0.631$} & \underline{$0.219$} \\
        \midrule
        \bottomrule
    \end{tabular}
    }
\end{table}

%% file: table/app_ablation.tex
\begin{table}[!ht]
    \caption{\textbf{Ablation Study of \thename on \celeba}. }
    \vskip-3mm
    \label{tab:app_ablation}
    \resizebox{\linewidth}{!}{
    \centering
        \begin{tabular}{l|rr|rr}
            \toprule
            {} & \multicolumn{2}{r|}{Male-to-Female} & \multicolumn{2}{r}{Female-to-Male} \\
            {} & FID & KID ($\times100$) & FID & KID ($\times100$) \\
            \midrule
            \uvcgan   & $9.6$ & $0.68\pm0.07$ & $13.9$ & $0.91\pm0.08$ \\
            \thename  & $\mathbf{4.7}$ & $\mathbf{0.14 \pm 0.02}$ & $\mathbf{7.6}$ & $\mathbf{0.24 \pm 0.02}$ \\
            \midrule
            (a) w/o New Gen.  & $8.1$ & $0.53 \pm 0.07$ & $11.1$ & $0.64 \pm 0.07$ \\
            (b) w/o New Disc. & $5.5$ & $0.21 \pm 0.03$ & $8.6$ & $0.31 \pm 0.03$ \\
            (c) w/o New Tech. & $5.6$ & $0.25 \pm 0.04$ & $10.0$ & $0.51 \pm 0.05$ \\
            \bottomrule
        \end{tabular}
    }
    \vskip-3mm
\end{table}

%% file: src/results_faithfulness.tex
\subsection{Toward Better Faithfulness Measures}
\label{sec:better_faithfulness}

\input{table/results_consist}

Pixel-wise image similarity measures (such as $L_2$, PSNR, and SSIM) have been shown~\cite{zhang2018unreasonable} to be weakly correlated with the human perception of similarity. However, they currently are being used~\cite{zhao2022egsde} as a faithfulness metric in the area of unpaired I2I translation.
In this section, we explore using CLIP~\cite{radford2021learning}
and LPIPS~\cite{zhang2018unreasonable} scores as similarity measures because they have been shown to match human perception.
In addition, it may be more natural to build a similarity measure reusing Inception-v3 features that are already employed in the FID calculation.
This will avoid introducing new dependencies, parameter choices, and other sources of inconsistency.
Thus, as another faithfulness metric, we propose an \textit{Inception-v3 $L_2$ distance} (I-$L_2$) formed between the Inception-v3 features.

\autoref{tab:results_consist} compares the faithfulness scores according to CLIP, LPIPS, I-$L_2$, and pixel-wise $L_2$.
It indicates that \thename variants outperform \egsde variants on the perceptual metrics.
On the other hand, \egsde shows superiority in the pixel-wise scores.
At the same time, \uvcgan demonstrates better perceptual faithfulness compared to \thename.
Given that \thename has better realism than \uvcgan,
this likely demonstrates an example of a realism-faithfulness trade-off.
The trends of the proposed I-$L_2$ roughly match those of the more complex CLIP and LPIPS measures, illustrating the feasibility of an I-$L_2$ as a simpler similarity measure.

Fundamentally, one still may wonder if any of these metrics are ``proper'' measurements of faithfulness?
When measuring faithfulness, one should differentiate between the \textit{domain-contrastive}
features (changing between domains) and \textit{domain-consistent} features (expected to be preserved).
For example, in the \celebahq dataset, males tend to have shorter hair compared to females.
Thus, we expect a Male-to-Female I2I algorithm will increase hair length on average
(domain-contrastive feature).
On the other hand, image background, skin color, facial expression, and facial orientation
are approximately the same in both domains.
Therefore, an I2I algorithm is expected to preserve these features (domain-consistent features).
We argue a proper faithfulness metric should pay the most attention
to domain-consistent features
and be indifferent to the domain-contrastive features.

\begin{figure}[ht]
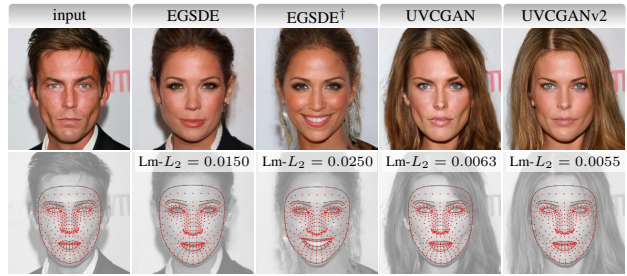

    \centering
    \tikzsetnextfilename{face_landmark}
    \resizebox{\linewidth}{!}{
    \begin{tikzpicture}
        \input{fig/grid}
        \def\folder{fig/face_landmarks}
        \def\index{1}
        
        \def\width{.12\textwidth}
        \def\expand{1.02}

        \node[inner sep=0, fill=white] at (0, 0) {
            \begin{tikzpicture}
                \foreach \name/\label [count=\col] in {input/input, 
                                                       egsde1/\egsde, 
                                                       egsde2/\egsdeDG,
                                                       uvcgan1/\uvcgan,
                                                       uvcgan/\thename} {
                    \def\fname{\folder/\name_\index.jpg}
                    \coordinate (C) at (\width * \expand * \col, 0);
        		  \node[inner sep=0] (image_\col) at (C) {\includegraphics[width=\width]{\fname}};
                    \tikzmath{coordinate \C;\C = (image_\col.north east) - (image_\col.south west);}
                    \node[header, minimum width=\Cx, minimum height=.2 * \width, anchor=south west] at (image_\col.north west) {\label};
                }
                \foreach \name/\label/\ll [count=\col] in {input/input/0, 
                                                           egsde1/\egsde/0.0150, 
                                                           egsde2/\egsdeDG/0.0250, 
                                                           uvcgan1/\uvcgan/0.0063,
                                                           uvcgan/\thename/0.0055} {
                    \def\fname{\folder/\name_\index_lm.jpg}
                    \coordinate (C) at (\width * \expand * \col, -\width * \expand);
        		  \node[inner sep=0] (image_\col) at (C) {\includegraphics[width=\width]{\fname}};
                    \ifthenelse{\col > 1}{
                        \node[fill=white, fill opacity=.7, text opacity=1, inner sep=2pt, anchor=north east] at (image_\col.north east) {\scriptsize Lm-$L_2 =\ll$};
                    }{}
                }
            \end{tikzpicture}
        };
    \end{tikzpicture}
    }
    \caption{Examples of face landmark distance (Lm-$L_2$) between an input and a translated image. 
             (\autoref{sec:Faithfulness})}
    \label{fig:face_landmark}
\end{figure}

As an initial attempt, we propose
an approximation to such a metric: similarity of face landmarks~\cite{facelm_creusot20103d,facelm_Burgos-Artizzu_2013_ICCV,facelm_mediapipe}.
Face landmarks are key points on a human face, capturing information about its
position, orientation, structure, expression, etc. That is, capturing those features expected to be largely preserved during the Male-to-Female translation.
Therefore, one may consider forming an Euclidean distance (Lm-$L_2$) between the landmark
locations of the source and translated images (\autoref{fig:face_landmark}). 
Better faithfulness metrics also will consider other domain-consistent features. This development is left for future work.

%% file: table/results_consist.tex
\begin{table}[t]
    \caption{\textbf{Faithfulness scores.}}
    \vskip-3mm
    \label{tab:results_consist}
    \centering
    \resizebox{.90\linewidth}{!}{
    \begin{tabular}{l|rrrr}
        \toprule
        {} & \multicolumn{4}{c}{Male-to-Female (CelebA-HQ)}  \\
        {} & CLIP$\uparrow$ & LPIPS$\downarrow$ & I-$L_2$$\downarrow$ & $L_2$$\downarrow$ \\ 
        \midrule
        \egsde   & $0.637$ & $0.116$ & $13.93$ & $\mathbf{43.41}$ \\
        \egsdeDG & $0.600$ & $0.131$ & $15.22$ & $54.50$ \\
        \uvcgan & $\mathbf{0.734}$ & $\mathbf{0.089}$ & $\mathbf{13.12}$ & $54.08$ \\
        \midrule
        \thename   & \underline{$0.723$} & $0.101$ & \underline{$13.36$} & $64.23$ \\
        \thename-C & $0.720$ & \underline{$0.093$} & $13.47$ & \underline{$47.94$} \\
        \midrule
        \midrule
        {} & \multicolumn{4}{c}{Cat-to-Dog} \\
        {} & CLIP$\uparrow$ & LPIPS$\downarrow$ & I-$L_2$$\downarrow$ & $L_2$$\downarrow$ \\ 
        \midrule
        \egsde   & $0.762$ & $0.165$ & $16.65$ & $\mathbf{49.20}$ \\
        \egsdeDG & $0.749$ & $0.181$ & $16.74$ & $62.44$ \\
        \uvcgan
            & $\mathbf{0.822}$ & $\mathbf{0.126}$ & $16.44$ & \underline{$54.45$} \\
        \midrule
        \thename   & $0.802$ & $0.150$ & $\mathbf{16.31}$ & $77.94$ \\
        \thename-C
            & \underline{$0.809$} & \underline{$0.140$}
            & \underline{$16.43$} & $56.83$ \\
        \midrule
        \midrule
        {} & \multicolumn{4}{c}{Wild-to-Dog} \\
        {} & CLIP$\uparrow$ & LPIPS$\downarrow$ & I-$L_2$$\downarrow$ & $L_2$$\downarrow$ \\ 
        \midrule
        \egsde   & $0.672$ & $0.176$ & $15.18$ & $\mathbf{56.65}$ \\
        \egsdeDG & $0.649$ & $0.192$ & $15.41$ & $68.46$ \\
        \uvcgan  & $\mathbf{0.765}$ & $\mathbf{0.124}$ & $15.14$ & $61.05$ \\
        \midrule
        \thename   & $0.720$ & $0.150$ & $\mathbf{14.75}$ & $81.75$ \\
        \thename-C & \underline{$0.740$} & \underline{$0.136$} & \underline{$14.81$} & \underline{$58.77$} \\
        \midrule
        \bottomrule
    \end{tabular}
    }
    \vskip-3mm
\end{table}

%% file: src/conclusions.tex
\section{Conclusion}

This work revisits the \uvcgan model, built upon the classic \cyclegan architecture.
It demonstrates that source-driven style modulation and batch-statistics-aware discriminator are effective techniques to improve the model performance.
Our \thename has been extensively benchmarked on four datasets with nine translation directions.
Results show our model can achieve superior performance in terms of FID scores.
At the same time, we note the absence of a proper faithfulness metric in the area of unpaired
I2I translation and general inconsistency of the evaluation procedures. As such, developing such a measure and uniformizing evaluation will be highly beneficial for future unpaired I2I methods development.

%% file: src/app_data.tex
\section{Datasets}
\label{sec:DatasetDetail}

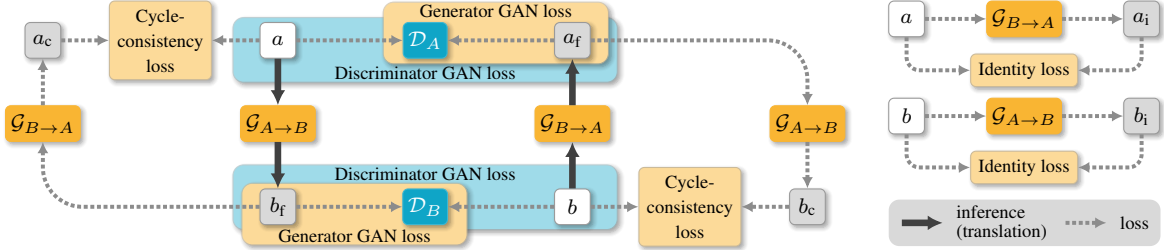
\begin{figure*}[t]
    \centering
        \tikzexternaldisable
        \tikzsetnextfilename{cyclegan}%
        \begin{tikzpicture}
            \node[inner sep=1pt, fill=white] at (0, 0) {\input{fig/cyclegan_2}};
        \end{tikzpicture}
    \caption{
        \textbf{\cyclegan framework.} 
        The \cyclegan~\cite{zhu2017unpaired} consists of two pairs of GANs, $\paren{\gab, \db}$ and $\paren{\gba, \da}$. 
        The discriminators try to distinguish translations from real images, 
        while the generators (or translators) seek to produce realistic translations that are also consistent with the input. 
        The consistency is enforced by the cycle-consistency loss and (optional) identity loss. 
        Here, we use $a$ to denote an image from domain $A$, $b$ as an image from domain $B$,
        $\fake{(*)}$ is a fake image (a translation), 
        $\reco{(*)}$ notes a cyclic reconstruction, 
        and $\iden{(*)}$ represents an identity reconstruction
        (when identity losses are used, $\left.\gab\right|_B$ and $\left.\gba\right|_A$ are encouraged to be identity maps).
    }
    \label{fig:cyclegan}
\end{figure*}

In this section, we provide additional details about the datasets used in the main paper.

\paragraph{\celeba~\cite{liu2015deep}.} The datasets for the \glasses Removal and \malefemale tasks are derived from the original \celeba dataset.
For a fair comparison with older models~\cite{torbunov2023uvcgan}, we used the pre-processed versions of
\glasses Removal and \malefemale datasets provided by \council~\cite{nizan2020breaking}.
The \celeba dataset is made of images of size $178 \times 218$ pixels. The train split of the \glasses Removal
task contains about 11K images with glasses and 152K without. The \malefemale dataset has about 68K males and 95K
females. The test parts of the \glasses Removal and \malefemale datasets contain about 3K images with glasses
and 37K images without glasses, and 16K males and 24K females respectively.

\paragraph{\animeselfie~\cite{kim2019u}.} The training split of the \animeselfie dataset contains 3400 Selfie images and 3400 Anime images.
The test part contains just 100 samples from each domain. All images of this dataset have a size of $256 \times 256$ pixels.

\paragraph{\celebahq~\cite{karras2017progressive}.}
The \celebahq dataset has around 10K images of males and about 18K images of females in the train split. The test split
contains 1000 male and 1000 female images. The size of a \celebahq image is $1024 \times 1024$ pixels.

\paragraph{\afhq~\cite{choi2020stargan}.}
The \afhq dataset has around 5.2K cat, 4.7K dog, and 4.7K wildlife images in the train split, and
500 images of each in the test split. The images of the \afhq dataset have a size of $512 \times 512$ pixels.
There are two versions of the \afhq dataset provided by StarGANv2~\cite{choi2020stargan}.
We use version 1 to be consistent with previous models.

%% file: fig/cyclegan_2.tex
    \begin{tikzpicture}[font=\fontsize{8}{8}\selectfont]
        \tikzstyle{block} = [
            rectangle, 
            rounded corners=2, 
            blur shadow={
                shadow blur steps=5, 
                shadow xshift=1pt, 
                shadow yshift=-1pt},
            align=center,
            draw=black!40, 
        	fill=black!15,
        ]
    
        \def\xs{150pt}
        \def\ys{160pt}
        \def\lw{1pt}
        
        \begin{scope}[x=\xs, y=\ys]
            \tikzstyle{data} = [block, minimum size=.09 * \xs]
            
            \tikzstyle{gen} = [block, minimum size=.09 * \xs, text=black, draw=or!100!black!80, fill=or!80, inner sep=2pt]
            \tikzstyle{dis} = [block, minimum size=.09 * \xs, text=white, draw=b2!100!black!80, fill=b2, inner sep=2pt]
            
            \tikzstyle{loss} = [block, inner sep=3pt, execute at begin node=\setlength{\baselineskip}{3ex}]
            \tikzstyle{genloss} = [loss, draw=or!50!black!40, fill=or!40, font=\fontsize{7}{7}\selectfont]
            \tikzstyle{disloss} = [loss, draw=b2!50!black!40, fill=b2!40, font=\fontsize{7}{7}\selectfont]
            
            \tikzstyle{flow} = [-{Latex[length=2mm, width=2mm]}, rounded corners=5pt, line width=2.5*\lw, draw=black!75]
            \tikzstyle{flowloss} = [-{Latex[length=1.5mm, width=1.5mm]}, rounded corners=5pt, line width=1.5 * \lw, densely dotted, draw=black!45]
            \tikzstyle{flowloss} = [-{Latex[length=1.5mm, width=1.5mm]}, rounded corners=5pt, line width=1.5 * \lw, densely dotted, draw=black!45]
            \tikzstyle{textnode} = [inner sep=1pt, font=\fontsize{7}{7}\selectfont]
    
            \def\xshift{.5 * \xs}
            \def\yshift{0.11 * \ys}
            
            \node[data, fill=white] (A) at (0, 0) {$a$};
            \node[gen, anchor=north] (GabA) at ([yshift=-\yshift]A.south) {$\gab$};
            \node[data, anchor=north] (FB) at ([yshift=-\yshift]GabA.south) {$\fake{b}$};
            \node[data, anchor=east] (CA) at ([xshift=-\xshift]A.west) {$\reco{a}$};
            \node[gen, anchor=north] (GbaA) at ([yshift=-\yshift]CA.south) {$\gba$};
            \draw[flow] (A) -- (GabA);
            \draw[flow] (GabA) -- (FB);
            
            \node[genloss, align=center] (CycleLossA) at ($(A)!.5!(CA)$) {Cycle-\\consistency\\loss};
            \draw[flowloss, rounded corners=15] (FB) -| (GbaA);
            \draw[flowloss] (GbaA) -- (CA);
            \draw[flowloss] (A) -- (CycleLossA);
            \draw[flowloss] (CA) -- (CycleLossA);

            
            \node[data, anchor=west, fill=white] (B) at ([xshift=1.3*\xshift]FB.east) {$b$};
            \node[gen, anchor=south] (GbaB) at ([yshift=\yshift]B.north) {$\gba$};
            \node[data, anchor=south] (FA) at ([yshift=\yshift]GbaB.north) {$\fake{a}$};
            \node[data, anchor=west] (CB) at ([xshift=\xshift]B.east) {$\reco{b}$};
            \node[gen, anchor=south] (GabB) at ([yshift=\yshift]CB.north) {$\gab$};

            \draw[flow] (B) -- (GbaB);
            \draw[flow] (GbaB) -- (FA);
            
            \node[genloss] (CycleLossB) at ($(B)!.5!(CB)$) {Cycle-\\consistency\\loss};
            \draw[flowloss, rounded corners=15] (FA) -| (GabB);
            \draw[flowloss] (GabB) -- (CB);
            \draw[flowloss] (B) -- (CycleLossB);
            \draw[flowloss] (CB) -- (CycleLossB);
            

            \node[dis] (DA) at ($(A)!.5!(FA)$) {$\da$};
            \node[dis] (DB) at ($(FB)!.5!(B)$) {$\db$};
            \draw[flowloss] (A) -- (DA);
            \draw[flowloss] (FA) -- (DA);
            \draw[flowloss] (B) -- (DB);
            \draw[flowloss] (FB) -- (DB);

            \node[data, anchor=north west, fill=white] (AA) at ([xshift=1.5 * \xshift]FA.east |-CycleLossA.north) {$a$};
            \node[data, anchor=west] (AI) at ([xshift=1 * \xshift]AA.east) {$\iden{a}$};
            \node[gen] (GbaI) at ($(AA)!.5!(AI)$) {$\gba$};
            \draw[flowloss] (AA) -- (GbaI);
            \draw[flowloss] (GbaI) -- (AI);
            \node[genloss, anchor=north] (AIdtLoss) at ([yshift=-.35 * \yshift]GbaI.south) {Identity loss};
            \draw[flowloss] (AA) |- (AIdtLoss);
            \draw[flowloss] (AI) |- (AIdtLoss);
            
            \node[data, anchor=north, fill=white] (BB) at ([yshift=-1.3 * \yshift]AA.south) {$b$};
            \node[data, anchor=west] (BI) at (AI.west |- BB) {$\iden{b}$};
            \node[gen] (GabI) at ($(BB)!.5!(BI)$) {$\gab$};
            
            \draw[flowloss] (BB) -- (GabI);
            \draw[flowloss] (GabI) -- (BI);
            \node[genloss, anchor=north] (BIdtLoss) at ([yshift=-.35 * \yshift]GabI.south) {Identity loss};
            \draw[flowloss] (BB) |- (BIdtLoss);
            \draw[flowloss] (BI) |- (BIdtLoss);

            \tikzmath{coordinate \L; \L=(AI.east)-(AA.west);}
            \node[minimum height=\yshift, minimum width=\Lx, rounded corners=4, fill=black!15, anchor=south west] (LegendBG) at (BB.west |- CycleLossB.south) {};
            \coordinate (TI) at ($(LegendBG.west)!.05!(LegendBG.east)$);
            \coordinate (HI) at ($(LegendBG.west)!.20!(LegendBG.east)$);
            \draw[flow] (TI) -- (HI);
            \node[textnode, anchor=west, align=left] (legendInf) at ([xshift=4pt]HI.east) {inference\\(translation)};
            \coordinate (TL) at ($(LegendBG.west)!.65!(LegendBG.east)$);
            \coordinate (HL) at ($(LegendBG.west)!.80!(LegendBG.east)$);
            \draw[flowloss] (TL) -- (HL);
            \node[textnode, anchor=west] (legendLoss) at ([xshift=4pt]HL.east) {loss};

            \tikzmath{coordinate \C;\C=(FA.east |- DA.north)-(A.west |- DA.south);}
            \def\dxsep{.58 * \Cx}
            \def\dysep{.88 * \Cy}
            \def\gxsep{.34 * \Cx}
            \def\gysep{.84 * \Cy}
            \begin{pgfonlayer}{back}
                \def\op{.6}
                \node[disloss, inner xsep=\dxsep, inner ysep=\dysep, rounded corners=4, draw opacity=\op] (DALoss) at ($(DA.north)!.76!(DA.south)$) {};
                \node[disloss, inner xsep=\dxsep, inner ysep=\dysep, rounded corners=4, draw opacity=\op] (DBLoss) at ($(DB.south)!.76!(DB.north)$)  {};
                \coordinate (V) at ($(FA)!.52!(DA)$);
                \coordinate (H) at ($(DA.south)!.70!(DA.north)$);
                \node[genloss, inner xsep=\gxsep, inner ysep=\gysep, rounded corners=4, draw opacity=\op] (GabGANLoss) at (V |- H) {};
                \coordinate (V) at ($(FB)!.52!(DB)$);
                \coordinate (H) at ($(DB.north)!.70!(DB.south)$);
                \node[genloss, inner xsep=\gxsep, inner ysep=\gysep, rounded corners=4, draw opacity=\op] (GbaGANLoss) at (V |- H) {};
                \node[textnode, anchor=south] at (DALoss.south) {Discriminator GAN loss};
                \node[textnode, anchor=north] at (DBLoss.north) {Discriminator GAN loss};
                \node[textnode, anchor=north] at (GabGANLoss.north) {Generator GAN loss};
                \node[textnode, anchor=south] at (GbaGANLoss.south) {Generator GAN loss};
            \end{pgfonlayer}
            
            
        \end{scope}    
    \end{tikzpicture}

%% file: src/app_train.tex
\section{Training Details}
\label{sec:TrainingDetails}

In this section, we expand on the generator pre-training, I2I translation training, HP optimization setup, the final model configurations, and the ablation study of \thename.

\paragraph{Generator Pre-Training.}
The pre-training of the generators was performed in a BERT-like fashion~\cite{devlin2018bert} on an image inpainting pretext task.

To construct the image inpainting task,
input images of size $256 \times 256$ pixels are tiled into a grid of patches of $32 \times 32$ pixels.
Then, each patch is randomly masked with a probability of $40\%$. The masking is performed by zeroing out
pixel values. The generator is tasked to recover the original unmasked image from a masked one.

For the pre-training, we use the AdamW optimizer together with a cosine learning rate annealing (with
restarts). We set the initial learning rate to $5 \times 10^{-3} \times (\text{batch size} / 512)$,
and the weight decay factor to $0.05$. The scheduler completes 5 annealing cycles during the
pre-training.

We apply several data augmentations, such as random rotation
($\pm 10$ degree), random horizontal flip ($p = 0.5$), and color jitter ($\pm 0.2$ shift in brightness,
contrast, saturation, and hue). 

We pre-train the generators for 500 epochs for \celebahq and \afhq. Due to the small size of
the \anime dataset, we run the pre-training for 2500 epochs. On the contrary, due to the large size
of the \celeba dataset, we run the pre-training for 500 epochs, but limit the number of samples
per epoch to $32,768$. All the pre-trainings are performed with a batch size of 64.

\paragraph{Image Translation Training.}
We train the unpaired I2I translation models by closely following the procedure
of \uvcgan~\cite{torbunov2023uvcgan}. We use the Adam optimizer without weight decay.
The training is performed for 1 million iterations.
We experiment with using either a constant learning rate or applying a linear scheduler.
If the linear scheduler is used, then the learning rate is maintained constant for the first
500K iterations, and then linearly annealed to zero during the subsequent 500K iterations.

We keep the batch size equal to 1 during the training. For consistency with
\progan~\cite{karras2017progressive} we keep the sizes of the image caches at 3.
This size effectively provides the batch head with 4 samples to estimate the batch statistics.
To stabilize the generator, we apply an exponential weight averaging to the generator with a momentum of
$0.9999$.

\paragraph{Hyperparameter Exploration.}
When performing the final training, we explored the following grid of hyperparameters:
\begin{itemize}
    \item Magnitude of the cycle-consistency loss $\lambda_\text{cyc}$: $\{ 5, 10\}$.
    \item Magnitude of the zero-centered gradient penalty $\lambda_\text{GP}$ : $\{ 0.001, 0.01, 1 \}$.
    \item Batch Head type: Batch Normalization (BN) vs Batch Standard Deviation (BSD).
    \item Generator's and Discriminator's learning rates:
        \begin{enumerate}
            \item Equal learning rates of $1 \times 10^{-4}$.
            \item Unequal learning rates, with the learning rate of the discriminator of $1 \times 10^{-4}$
                and the learning rate of the generator of $5 \times 10^{-5}$.
        \end{enumerate}
\end{itemize}

The hyperparameter explorations were performed while keeping the magnitude of the consistency loss $\lambda_\text{consist}$
equal to zero.
The grid of hyperparameters above was suggested by the previous rough HP exploration.

For the \afhq \catdog, \wilddog, and \celebahq \malefemale datasets, we have run a second
hyperparameter exploration, studying the effect of the magnitude of the consistency
loss $\lambda_\text{consist}$ on the I2I performance. We have tried the following
values of $\lambda_\text{consist}$: $\{ 0.01, 0.1, 0.2, 0.4, 0.6, 0.8, 1.0 \}$.

\subsection{Final Configurations}

\input{table/app_train_configs}

For all the translation tasks, \thename achieves the best performance when the learning rate scheduler is not used.
\autoref{tab:app_train_configs} summarizes the final hyperparameter configurations (generator's learning rate, magnitudes of the gradient penalty and the cycle consistency loss, and the choice of batch head) that
provide the best performance per translation task.

Generally, the high-quality datasets (\celebahq and \afhq) favor stronger values of the gradient penalty term $\lambda_\text{GP} = 1$, compared to the lower-resolution datasets (\animeselfie and \celeba), favoring $\lambda_\text{GP} = 0.01$. Other
patterns of hyperparameters can be observed in the table. However, their impact
on the model's performance is relatively small compared to $\lambda_\text{GP}$.

We should note, that due to the instability associated with GANs training, some of the best values in \autoref{tab:app_train_configs} may be due to random fluctuations.

\subsection{Ablation Study}

\input{table/app_ablation_full}

In this section we provide a more detailed ablation study of \thename.
\autoref{tab:app_ablation_full} summarizes \thename ablation results on the \malefemale
translation of \celeba. To produce this table we start with the final \thename configuration and make one of following modifications separately:
(a) disable style modulation in the generator;
(b) disable batch head of the discriminator;
(c) use \uvcgan training setup: linear scheduler, full GP, and disable generator averaging;
(d) disable spectral normalization (SN);
(e) add linear schedule;
(f) remove exponential averaging of the generator weights.

\autoref{tab:app_ablation_full} shows, that the removal of the spectral normalization (d) decreases the model performance, but the effect is negligible. This indicates that the Spectral Normalization might be redundant and can be removed.

Each of the items (e) and (f) of \autoref{tab:app_ablation_full} may suggest that either the scheduler or exponential averaging of the generator weights is detrimental to the
model's performance. However, this is not the case, since the effects of (d) and (e) are entangled and mutually balancing. Individual modifications to (d) or (e)
destroy the balance and produce large changes in the model's performance. These changes are not indicative of the effects of the joint modifications as shown by (c).

%% file: table/app_train_configs.tex
\begin{table}[!ht]
    \caption{\textbf{Best Training Configurations.}}
    \vskip-3mm
    \label{tab:app_train_configs}
    \resizebox{\linewidth}{!}{
    \centering
        \begin{tabular}{l|r|r|r|c}
            \toprule
            Dataset
                & $\text{lr}_\text{gen}$ & $\lambda_\text{GP}$
                & $\lambda_\text{cyc}$ & B. Head \\
            \midrule
            Anime to Selfie & $5 \times 10^{-5}$ & $0.01$ & $10$ & BN \\
            Male to Female  & $1 \times 10^{-4}$ & $0.01$ & $10$ & BSD \\
            Glasses Removal & $5 \times 10^{-5}$ & $0.01$ & $10$ & BSD \\
            HQ Male to Female & $1 \times 10^{-4}$ & $1.0$ & $5$ & BSD \\
            Cat to Dog      & $1 \times 10^{-4}$ & $1.0$  & $5$  & BN \\
            Wild to Dog     & $1 \times 10^{-4}$ & $1.0$  & $5$  & BN \\
            Cat to Wild     & $5 \times 10^{-5}$ & $1.0$  & $5$  & BN \\
            \bottomrule
        \end{tabular}
    }
\end{table}

%% file: table/app_ablation_full.tex
\begin{table}[!ht]
    \caption{\textbf{Ablation Study of \thename on \celeba}. "$+$" indicates that an option is added to the final \thename configuration. "$-$" indicates that an option is removed from \thename.}
    \vskip-3mm
    \label{tab:app_ablation_full}
    \resizebox{\linewidth}{!}{
    \centering
        \begin{tabular}{l|rr|rr}
            \toprule
            {} & \multicolumn{2}{r|}{Male-to-Female} & \multicolumn{2}{r}{Female-to-Male} \\
            {} & FID & KID ($\times100$) & FID & KID ($\times100$) \\
            \midrule
            \uvcgan   & $9.6$ & $0.68\pm0.07$ & $13.9$ & $0.91\pm0.08$ \\
            \thename  & $\mathbf{4.7}$ & $\mathbf{0.14 \pm 0.02}$ & $\mathbf{7.6}$ & $\mathbf{0.24 \pm 0.02}$ \\
            \midrule
            (a) $-$ New Gen.  & $8.1$ & $0.53 \pm 0.07$ & $11.1$ & $0.64 \pm 0.07$ \\
            (b) $-$ New Disc. & $5.5$ & $0.21 \pm 0.03$ & $8.6$ & $0.31 \pm 0.03$ \\
            (c) $-$ New Tech. & $5.6$ & $0.25 \pm 0.04$ & $10.0$ & $0.51 \pm 0.05$ \\
            (d) $-$ SN         & $4.7$ & $0.14 \pm 0.02$ & $7.7$ & $0.25 \pm 0.03$ \\
            \midrule
            (e) $+$ Sched      & $6.3$ & $0.35 \pm 0.05$ & $9.5$ & $0.47 \pm 0.05$ \\
            (f) $-$ Avg.       & $9.8$ & $0.71 \pm 0.05$ & $14.2$ & $0.91 \pm 0.07$ \\
            \bottomrule
        \end{tabular}
    }
\end{table}

%% file: src/app_eval_inconsistency.tex
\section{Remarks on Metric Evaluation Consistency}
\label{sec:RemarksMetricEvalConsistency}

Inconsistency of unpaired I2I evaluation procedures is a widespread problem.
For example, some works (e.g.~\cite{kim2019u}) roll out their own FID evaluation code~\cite{ganmetricsmean} and report the so-called "mean" FID and KID scores, where ``mean'' means a weighted average of
the actual FID/KID scores and various other metrics. Some other works~\cite{parmar2022aliased}
choose different image resizing algorithms, creating a noticeable discrepancy in the reported FID scores.

Another source of FID/KID score inconsistency is the difference in the evaluation protocols. For instance,
works like~\cite{torbunov2023uvcgan} prefer to evaluate FID scores only on images of the test split, yet others~\cite{choi2020stargan} evaluate FID scores between  translated test images and target
images of the train split. Likewise, there is a difference in whether any pre-processing is used in the FID/KID evaluation.
For example, one can evaluate FID scores between translated images with the pre-processing and target images
without pre-processing~\cite{park2020contrastive}, or one can apply the pre-processing step to both translated and
target images~\cite{zhao2022egsde}. Moreover, the pre-processing procedures may differ between different works.

To uniformize FID/KID evaluation procedures, we propose a consistent evaluation protocol in \autoref{sec:consistent_eval}.

%% file: src/app_fidelity.tex
\section{Metrics of Faithfulness to the Source}
\label{sec:Faithfulness}

\input{fig/grid_consistency_suppl}

In this section, we provide more discussion on the two faithfulness metrics we proposed: 
Inception v3 $L_2$ (I-$L_2$) and Landmark $L_2$ (Lm-$L_2$).

\vskip2mm
\noindent\textbf{Inception v3 $L_2$ Distance:} Here we present more examples to illustrate
that I-$L_2$ (defined as the $L_2$ distance
between the latent Inception-v3 features) may be a more appropriate measurement of faithfulness than pixel-level $L_2$.
In \autoref{tab:comparingTransWithL2}, we select translations according to two types of criteria.
Denote a translation produced by \egsde as $E$ and that by \thename as $U$.
\[
\begin{aligned}
&\text{\textbf{Type 1}: I-$L_2(E) \approx $ I-$L_2(U)$ and $L_2(U) - L_2(E) > 15$.}\\
&\text{\textbf{Type 2}: I-$L_2(E) - $ I-$L_2(U) > 3$.}
\end{aligned}
\]
\textbf{Type~1} is designed to show what contributes to lower pixel-wise $L_2$ while I-$L_2$ scores are similar. 
\textbf{Type~2} selects examples with large I-$L_2$ difference and helps readers to judge if I-$L_2$ correlates with their own judgment on the similarity to the source (i.e.~which pairs look more like siblings.)

We list eight categories of perceived image faithfulness (PIF) in the legend such as background, bone structure, facial expression, and so on.  
For each translated image, we indicate which categories it outperforms that generated by the other model.
For example, for the input in Type 1 row 1 left, \egsde preserves the hairstyle and color (PIF=7) better than \thename, 
but the \thename translation maintains a sharper background~(1), preserves the bone structure~(2) and expression~(3) better, 
and exhibits more similarity in apparent age~(4). 

Type-1 examples suggest that pixel-level $L_2$ is an inappropriate measurement for semantic consistency as \thename translations manage to capture characterizing features (such as a bone structure) even with high pixel-level $L_2$. 
In fact, the high pixel-level $L_2$ of \thename translations is often a result of benign modifications such as the elongation of dark hair on a light background (e.g.~Type 1 row 5 right) or a slight overall shift to a warmer hue (e.g.~Type 1 row 2 right). 

On the contrary, the Type-2 examples suggest that I-$L_2$, the $L_2$ on Inception latent features, might be a better measurement of semantic consistency. 
While \egsde fails to maintain features such as background, facial expression (neural v.s.~smile), eye movement, and prominent bone structure and produces over-generalized translations, \thename translations with significantly lower I-$L_2$ manage to preserve those features and appear more individualized.

These examples illustrate that the pixel-wise $L_2$ faithfulness metric may be in poor agreement with a human judgment of image faithfulness. They also point to a possibility that the  I-$L_2$ distances, based on deep features of Inception-v3, may better capture the perceived image faithfulness. Such observations mirror the conclusion of~\cite{zhang2018unreasonable}, about the effectiveness of deep features as perceptual metrics.

However, we stress again, that the main purpose of this paper is to improve the performance of the classic \cyclegan architecture, not the development of better faithfulness metrics. 
While this section points to a possibility of I-$L_2$ being a better faithfulness metric, a full-scale investigation needs to be conducted to conclusively establish this. 
We leave such a study for future work.

\vskip2mm
\noindent\textbf{Landmark $L_2$ Distance:} We start with defining Lm-$L_2$.
Let $\mathcal{L}^{\textrm{i}} = \paren{l^{\textrm{i}}_1, \dots, l^{\textrm{i}}_N}$ 
be the landmarks of an input face and let 
$\mathcal{L}^{\textrm{t}} = \paren{l^{\textrm{t}}_1, \dots, l^{\textrm{t}}_N}$ 
be those of a translated face.
Lm-$L_2$ is defined as follows:
\[
    \textrm{Lm-}L_2\paren{\mathcal{L}^{\textrm{i}}, \mathcal{L}^{\textrm{t}}} = \frac{1}{N}\sum_{i=1}^{N}\left\Vert l^\textrm{i}_i - l^\textrm{t}_i \right\Vert_2.
\]
For this study, we calculate face landmarks with \textsf{mediapipe} (v0.10.1)~\cite{facelm_mediapipe} 
with $N=468$ and $l_i=(x, y, z)$ being the location of the landmark relative to the image.
From \autoref{tab:results_consist_ll2}, we can see that \thename is more faithful with respect to Lm-$L_2$.
Find more examples of face landmarks with their Lm-$L_2$ in \autoref{tab:grid_face_landmarks}.

\begin{table}[t]
    \caption{\textbf{Faithfulness by Landmark $L_2$ (Lm-$L_2$).}}
    \vskip-3mm
    \label{tab:results_consist_ll2}
    \centering
    \resizebox{.95\linewidth}{!}{
    \begin{tabular}{l|rrrrr}
        \toprule
                         & \egsde & \egsdeDG & \uvcgan & \thename & \thename-C \\ 
        \midrule
        Lm-$L_2$$\downarrow$ & $0.0188$ & $0.0227$ & $0.0129$ & $0.0152$ & $0.0163$ \\
        \bottomrule
    \end{tabular}
    }
\end{table}

\begin{table*}[ht]
    \centering
    \caption{\textbf{Face landmarks with Lm-$L_2$}}
    \vskip-3mm
    \tikzexternaldisable
    \tikzsetnextfilename{grid_face_landmarks}
    \input{fig/grid}
    \resizebox{\textwidth}{!}{
    \begin{tikzpicture}
        \node[inner sep=0, fill=white] at (0, 0) {
            \begin{tikzpicture}
                \node[inner sep=0] (G1) at (0, 0) {
                    \def\folder{fig/face_landmarks}
                    \def\width{.104\textwidth}
                    \def\expand{1.03}
                    \gridlm{0, 1, 2, 3, 4}
                           {input/input, 
                            egsde1/\egsde, 
                            egsde2/\egsdeDG,
                            uvcgan1/\uvcgan,
                            uvcgan/\thename,
                            uvcgan_const/\thename-C}
                };
                \node[inner sep=0, anchor=north] (G2) at ([yshift=-5pt]G1.south) {
                    \def\folder{fig/face_landmarks}
                    \def\width{.104\textwidth}
                    \def\expand{1.03}
                    \gridlm[5]{5, 6, 7, 8, 9}
                              {input/input, 
                              egsde1/\egsde, 
                              egsde2/\egsdeDG, 
                              uvcgan1/\uvcgan,
                              uvcgan/\thename,
                              uvcgan_const/\thename-C}
                };   
            \end{tikzpicture}
        };
    \end{tikzpicture}
    }
    \label{tab:grid_face_landmarks}
\end{table*}

%% file: fig/grid_consistency_suppl.tex
\tikzstyle{imagenode} = [inner sep=0, anchor=west]

\tikzstyle{textBG} = [
    inner sep=0, 
    minimum width=\width, 
    minimum height=\width, 
    text width=.9 * \width,
    anchor=west,
    align=left, 
    fill=black!15,
    font=\fontsize{6}{6}\selectfont
]

\tikzstyle{header} = [
    inner sep=0, 
    minimum width=\width, 
    minimum height=.25 * \width, 
    anchor=south,
    rounded corners=2,
    fill=black!30,
    path fading=south,
    font=\fontsize{8}{8}\selectfont,
]

\tikzstyle{index} = [
    inner sep=0, 
    rotate=90,
    anchor=south,
    rounded corners=2,
    fill=black!30,
    path fading=east,
    font=\fontsize{8}{8}\selectfont,
]

\def\folder{fig/grid_consist}
\def\width{.0935\textwidth}
\def\expand{1.03}
\def\comWidth{.2\textwidth}

\newcommand{\fillError}[5]{
    \pgfplotstablegetelem{#1}{l2_#2}\of\table;
    \pgfmathsetmacro\IL{\pgfplotsretval};
    \pgfplotstablegetelem{#1}{pixel_l2_#2}\of\table;
    \pgfmathsetmacro\PL{\pgfplotsretval};
    \pgfplotstablegetelem{#1}{#2}\of#5
    \coordinate (local_temp) at (\width * \expand * #4, -\width * \expand * #3);
    \node[textBG] (\row_#4) at (local_temp) {
        \setlength\tabcolsep{1pt}
        \renewcommand{\arraystretch}{1.2}
        \begin{tabular}{rcp{15pt}}    
            $L_2$ & $=$ & $\round[1]{\PL}$\\
            I-$L_2$ & $=$ & $\round[1]{\IL}$\\
            PIF & $=$ & \pgfplotsretval
        \end{tabular}
    };
}

\newcounter{dummy}

\newcommand{\gridSym}[4][]{
    \begin{tikzpicture}
        \pgfplotstableread[col sep=comma]{\folder/male2female_#2.csv}\table        
        \setcounter{dummy}{0}%
        \foreach \idx [count=\row] in {#3} {
            \stepcounter{dummy}
            \foreach \name/\col in {input/0, EGSDE1/-1, UVCGAN2wo/1}{
                \def\fname{\folder/male2female_#2_\name_\idx.png.jpg}
                \coordinate (local_temp) at (\width * \expand * \col, -\width * \expand * \row);
                \node[imagenode] (\row_\col) at (local_temp) {\includegraphics[width=\width]{\fname}};
            }
            \fillError{\idx}{EGSDE1}{\row}{-2}{#4}
            \fillError{\idx}{UVCGAN2wo}{\row}{2}{#4}
        }
        \node[header] (h_input) at (1_0.north) {input};
        \tikzmath{ coordinate \C; \C = (1_-1.north east) - (1_-2.south west); }
        \node[header, minimum width=\Cx, anchor=south west] at (1_-2.north west) {\egsde};
        \tikzmath{ coordinate \C; \C = (1_2.north east) - (1_1.south west); }
        \node[header, minimum width=\Cx, anchor=south west] at (1_1.north west) {\thename};
        \edef\num{\arabic{dummy}}
        \ifthenelse{\equal{#1}{}} 
                   {} 
                   {\node[index, minimum width=\num * \width * \expand - \width * \expand + \width, minimum height=.28 * \width, anchor=south east] at (1_-2.north west) {#1};}
    \end{tikzpicture}
}

\pgfplotstableread[col sep=&]{typeOneComment.tab}{\commentOne}

\pgfplotstableread[col sep=&]{typeTwoComment.tab}{\commentTwo}

\begin{table*}
    \centering
    \caption{\textbf{Comparing \egsde and \thename translations with $L_2$ and I-$L_2$ scores.}}
    \vskip-3mm
    \label{tab:comparingTransWithL2}
    \tikzexternaldisable
    \tikzsetnextfilename{app-comparison_type1_1}
    \begin{tikzpicture}
        \node[inner sep=2pt, fill=white] at (0, 0) {
            \begin{tikzpicture}
                \node[inner sep=0] (Type1left) at (0, 0) {\gridSym[Type 1]{type1}{0, ..., 6}{\commentOne}};
                \node[inner sep=0, anchor=west] (Type1right) at ([xshift=3pt]Type1left.east) {\gridSym{type1}{7, ..., 13}{\commentOne}};
                \node[inner sep=0, anchor=north] (Type2left) at ([yshift=-5pt]Type1left.south) {\gridSym[Type 2]{type2}{0, ..., 2}{\commentTwo}};
                \node[inner sep=0, anchor=north] (Type2right) at ([yshift=-5pt]Type1right.south) {\gridSym{type2}{3, ..., 5}{\commentTwo}};
        
                \node[inner sep=5pt, 
                      draw=black!40,
                      thick,
                      anchor=north east, 
                      fill=black!20,
                      blur shadow={shadow blur steps=5},
                      rounded corners] at ([xshift=-1pt, yshift=-5pt]Type2right.south east) 
                {
                    \begin{tikzpicture}
                        \def\lwidth{1.28 * \width}
                        \tikzstyle{legendCell} = [inner sep=0, align=left, anchor=north, font=\fontsize{9}{9}\selectfont]
                        \foreach \cat [count=\idx] in {background, bone structure, expression, apparent age, eye color, eyebrows, hair color, hair texture} {
                            \node[legendCell, 
                                 ] (1_\idx) at (\idx * \lwidth, 0) {\idx.~\cat};
                        }
                        \node[legendCell, anchor=south west] at ([xshift=-4pt, yshift=4pt]1_1.north west) {Categories of perceived image faithfulness (PIF):};
                    \end{tikzpicture}
                };
            \end{tikzpicture}
        };
    \end{tikzpicture}
\end{table*}

%% file: src/app_results_consistent.tex
\section{Toward Consistent FID Evaluation}
\label{sec:consistent_eval}

\input{table/app_results_hq}

The evaluation protocols used in the paper for \celebahq and \afhq are provided by \egsde~\cite{zhao2022egsde}.
Being ad-hoc, these protocols lack consistency and differ significantly
depending on the dataset.
A variety of different evaluation protocols makes the evaluation of the unpaired I2I methods rather quirky and error-prone.

As a step toward consistent FID evaluation,
we extend a more uniform evaluation protocol of \uvcgan~\cite{torbunov2023uvcgan}
toward the AFHQ and CelebA-HQ datasets.
The \thename evaluation results under this consistent evaluation protocol are shown
in \autoref{tab:app_results_hq}.
The consistent evaluation protocol uses only test splits (or validation splits if the test ones are not available) of each dataset to assess the quality of image translation.

The evaluation protocol begins with pre-processing all the 
 datasets in a consistent manner. The pre-processing step resizes images from their
original size down to $256 \times 256$ pixels (the same image size as is used for model training and inference). To avoid FID score inconsistencies created by aliasing artifacts~\cite{parmar2022aliased} we rely on the \texttt{Pillow} library~\cite{pillow} and Lanczos interpolation method.

Once the data pre-processing and image translation are done, the actual evaluation can begin. To perform the FID/KID score computation we use
a \texttt{torch-fidelity} package~\cite{obukhov2020torchfidelity}, which provides
a validated implementation of these metrics. The KID evaluation procedure depends on a free parameter -- the KID subset size. In this section, we choose the KID subset size of 100 for all the datasets.

The suggested evaluation protocol differs in a number of ways from the evaluation
protocols of the \afhq and \celebahq datasets of \egsde. It differs from the
ad-hoc \celebahq evaluation protocol~\cite{zhao2022egsde} because the latter compares FID scores between samples of validation and train splits, while the consistent version only uses validation split. The consistent evaluation protocol is also different from  the ad-hoc version of the \afhq one, which performs FID evaluation between translated images of size $256 \times 256$ and target images of size $512 \times 512$. The consistent protocol always uses pre-processed images of size $256 \times 256$.

\subsection{\celebahq Consistent Evaluation}
\label{sec:app_train_uvcgan_consist}

In this section, we elaborate on the \celebahq FID scores obtained with the
consistent protocol.
\autoref{tab:app_results_hq} shows the FID evaluation scores of \thename according to the 
consistent evaluation protocol. A few observations one can draw comparing the
results from this table to the ad-hoc evaluations, presented in Table 2 of the main paper.
In particular, there is a good agreement of the FID scores on the \afhq dataset,
since the ad-hoc evaluation procedure is very similar to the consistent one.
However, there is a large discrepancy in the FID scores on \celebahq.

Several factors contribute to the discrepancy of the FID scores on the \celebahq dataset. First,
the ad-hoc method evaluates the FID scores against the train part of the dataset, while the
consistent method uses validation parts.
This explains the large difference in the scale of the FID values~\cite{binkowski2018demystifying}.
Second, the consistent evaluation uses a Lanczos interpolation method,
less affected by aliasing artifacts.
Finally, the consistent version relies on a validated FID calculation code of \texttt{torch-fidelity},
while the ad-hoc method of \egsde uses a custom FID implementation.

Due to the large differences in the evaluation protocols on the \celebahq dataset,
it may be instructive to compare FID scores on \celebahq,
obtained with the consistent protocol.
\autoref{tab:app_results_uvcgan_consist}
compares the performance of \egsde, \uvcgan, and \thename under the consistent evaluation mode.
To a large extent, the behavior of the FID scores in \autoref{tab:app_results_uvcgan_consist}
matches the behavior of the FID scores of the ad-hoc protocol.
At the same time, according to the consistent evaluation protocol, \thename outperforms
\uvcgan by $15 - 20 \%$ in the FID scores.
This is in contrast to the ad-hoc evaluation procedure of Table 2 (the main part of the paper)
which does not reflect any difference in the performance.

%% file: table/app_results_hq.tex
\begin{table}[t]
    \caption{\textbf{Consistent FID and KID scores.}}
    \vskip-3mm
    \label{tab:app_results_hq}
    \resizebox{\linewidth}{!}{
    \centering
        \begin{tabular}{l|rr|rr}
            \toprule
            {} & \multicolumn{2}{r|}{Female to Male}
               & \multicolumn{2}{r}{Male to Female} \\
            {} & FID$\downarrow$ & KID ($\times100$) & FID$\downarrow$ & KID ($\times100$) \\ 
            \midrule
            \thename   & $29.7$ & $0.41 \pm 0.18$ & $24.2$ & $0.20 \pm 0.15$ \\
            \midrule
            \midrule
            {} & \multicolumn{2}{r|}{Dog to Cat}
               & \multicolumn{2}{r}{Cat to Dog} \\
            {} & FID & KID ($\times100$) & FID & KID ($\times100$) \\
            \midrule
            \thename   & $24.8$ & $0.23 \pm 0.13$ & $44.2$ & $0.76 \pm 0.23$ \\
            \midrule
            \midrule
            {} & \multicolumn{2}{r|}{Dog to Wild}
               & \multicolumn{2}{r}{Wild to Dog} \\
            {} & FID & KID ($\times100$) & FID & KID ($\times100$) \\
            \midrule
            \thename   & $18.7$ & $0.15 \pm 0.14$ & $44.7$ & $0.68 \pm 0.23$ \\
            \midrule
            \midrule
            {} & \multicolumn{2}{r|}{Cat to Wild}
               & \multicolumn{2}{r}{Wild to Cat} \\
            {} & FID & KID ($\times100$) & FID & KID ($\times100$) \\
            \midrule
            \thename   & $12.1$ & $0.01 \pm 0.09$ & $21.2$ & $0.20 \pm 0.13$ \\
            \bottomrule
        \end{tabular}
    }
\end{table}

%% file: table/app_results_consitency_effect.tex
\begin{table}[t]
    \caption{\textbf{Model Performance versus magnitude of the pixel-wise consistency loss.}}
    \vskip-3mm
    \label{tab:app_results_lambda_consist}
    \centering
        \begin{tabular}{r|rr|rr}
            \toprule
             & \multicolumn{2}{r|}{Male-to-Female}
               & \multicolumn{2}{r}{Cat-to-Dog} \\
            $\lambda_\text{consist}$ & FID$\downarrow$ & $L_2\downarrow$ & FID$\downarrow$ & $L_2\downarrow$ \\ 
            \midrule
            $0$    & $24.2$ & $62.6$ & $44.2$ & $77.9$ \\
            $0.01$ & $24.9$ & $61.2$ & $44.5$ & $76.7$ \\
            $0.1$  & $24.8$ & $50.6$ & $45.7$ & $64.9$ \\
            $0.2$  & $25.1$ & $47.9$ & $51.8$ & $56.8$ \\
            $0.4$  & $27.3$ & $43.3$ & $59.1$ & $50.6$ \\
            $0.6$  & $29.7$ & $41.0$ & $71.3$ & $47.5$ \\
            $0.8$  & $32.0$ & $39.1$ & $77.0$ & $46.1$ \\
            $1.0$  & $33.1$ & $37.7$ & $81.2$ & $44.9$ \\
            \bottomrule
        \end{tabular}
\end{table}

%% file: table/app_uvcgan_consist.tex
\begin{table}[t]
    \caption{\textbf{Consistent \uvcgan evaluation on \celebahq.}}
    \vskip-3mm
    \label{tab:app_results_uvcgan_consist}
    \resizebox{\linewidth}{!}{
    \centering
        \begin{tabular}{l|rr|rr}
            \toprule
            {} & \multicolumn{2}{r|}{Female to Male}
               & \multicolumn{2}{r}{Male to Female} \\
            {} & FID$\downarrow$ & KID ($\times100$) & FID$\downarrow$ & KID ($\times100$) \\ 
            \midrule
            \egsde   & - & - & $68.3$ & $6.10 \pm 0.48$ \\
            \egsdeDG & - & - & $53.9$ & $4.31 \pm 0.35$ \\
            \uvcgan  & $34.6$ & $0.78 \pm 0.40$ & $30.0$ & $0.74 \pm 0.32$ \\
            \midrule
            \thename   & $\mathbf{29.7}$ & $\mathbf{0.41 \pm 0.18}$
                & $\mathbf{24.2}$ & $\mathbf{0.20 \pm 0.15}$ \\
            \bottomrule
        \end{tabular}
    }
\end{table}

%% file: src/app_figures.tex
\section{Additional Translation Samples}
\label{sec:additionalTransSamples}

\def\width{.104\textwidth}
\def\expand{1.03}

\def\folder{fig/grid_LQ}
\begin{table*}[p]
    \centering
    \caption{\textbf{Sample translations for \selfieanime on \anime.}}
    \label{fig:grid_LQ_selfieanime}
    \tikzexternaldisable
    \tikzsetnextfilename{grid_LQ_selfie2anime}
    \input{fig/grid}
    \begin{tikzpicture}
        \node[inner sep=0, fill=white] (G1) at (0, 0) {
            \gridcp{selfie2anime}
                   {0, ..., 8}
                   {input/input, 
                    CycleGAN/\cyclegan, 
                    U-GAT-IT/\ugatit, 
                    UVCGAN/\uvcgan, 
                    UVCGAN2/\thename}
        };
    \end{tikzpicture}
\end{table*}

\begin{table*}[p]
    \centering
    \caption{\textbf{Sample translations for \animeselfie on \anime.}}
    \label{fig:grid_LQ_animeselfie}
    \tikzexternaldisable
    \tikzsetnextfilename{grid_LQ_anime2selfie}
    \input{fig/grid}
    \begin{tikzpicture}
        \node[inner sep=0, fill=white] (G1) at (0, 0) {
            \gridcp{anime2selfie}
                   {0, ..., 8}
                   {input/input, 
                    CycleGAN/\cyclegan, 
                    U-GAT-IT/\ugatit, 
                    UVCGAN/\uvcgan, 
                    UVCGAN2/\thename}
        };
    \end{tikzpicture}
\end{table*}

\begin{table*}[p]
    \centering
    \caption{\textbf{Sample translations for \malefemale on \celeba.}}
    \label{fig:grid_LQ_malefemale}
    \tikzexternaldisable
    \tikzsetnextfilename{grid_LQ_male2female}
    \input{fig/grid}
    \begin{tikzpicture}
        \node[inner sep=0, fill=white] (G1) at (0, 0) {
            \gridcp{male2female}
                   {0, 1, 2, 3, 4, 11, 6, 7, 8}
                   {input/input, 
                    CycleGAN/\cyclegan, 
                    U-GAT-IT/\ugatit, 
                    UVCGAN/\uvcgan, 
                    UVCGAN2/\thename}
        };
    \end{tikzpicture}
\end{table*}

\begin{table*}[p]
    \centering
    \caption{\textbf{Sample translations for \femalemale on \celeba.}}
    \label{fig:grid_LQ_femalemale}
    \tikzexternaldisable
    \tikzsetnextfilename{grid_LQ_female2male}
    \input{fig/grid}
    \begin{tikzpicture}
        \node[inner sep=0, fill=white] (G1) at (0, 0) {
            \gridcp{female2male}
                   {10, 11, 2, 3, 4, 5, 6, 7, 8}
                   {input/input, 
                    CycleGAN/\cyclegan, 
                    U-GAT-IT/\ugatit, 
                    UVCGAN/\uvcgan, 
                    UVCGAN2/\thename}
        };
    \end{tikzpicture}
\end{table*}

\begin{table*}[p]
    \centering
    \caption{\textbf{Sample translations for \rmvGlasses on \celeba.}}
    \label{fig:grid_LQ_rmvGlasses}
    \tikzexternaldisable
    \tikzsetnextfilename{grid_LQ_rmvGlasses}
    \input{fig/grid}
    \begin{tikzpicture}
        \node[inner sep=0, fill=white] (G1) at (0, 0) {
            \gridcp{rmvGlasses}
                   {0, ..., 8}
                   {input/input, 
                    CycleGAN/\cyclegan, 
                    U-GAT-IT/\ugatit, 
                    UVCGAN/\uvcgan, 
                    UVCGAN2/\thename}
        };
    \end{tikzpicture}
\end{table*}

\begin{table*}[p]
    \centering
    \caption{\textbf{Sample translations for \addGlasses on \celeba.}}
    \label{fig:grid_LQ_addGlasses}
    \tikzexternaldisable
    \tikzsetnextfilename{grid_LQ_addGlasses}
    \input{fig/grid}
    \begin{tikzpicture}
        \node[inner sep=0, fill=white] (G1) at (0, 0) {
            \gridcp{addGlasses}
                   {0, ..., 8}
                   {input/input, 
                    CycleGAN/\cyclegan, 
                    U-GAT-IT/\ugatit, 
                    UVCGAN/\uvcgan, 
                    UVCGAN2/\thename}
        };
    \end{tikzpicture}
\end{table*}

\def\folder{fig/grid_HQ}

\begin{table*}[p]
    \centering
    \caption{\textbf{Sample translations for \catdog on \afhq.}}
    \label{fig:grid_HQ_catdog}
    \tikzexternaldisable
    \tikzsetnextfilename{grid_HQ_cat2dog}
    \input{fig/grid}
    \begin{tikzpicture}
        \node[inner sep=0, fill=white] (G1) at (0, 0) {
            \gridcp{cat2dog}
                   {9, ..., 17}
                   {input/input, 
                    EGSDE1/\egsde, 
                    EGSDE2/\egsdeDG,
                    UVCGAN1/\uvcgan,
                    UVCGAN2wo/\thename}
        };
    \end{tikzpicture}
\end{table*}

\begin{table*}[p]
    \centering
    \caption{\textbf{Sample translations for \wilddog on \afhq.}}
    \label{fig:grid_HQ_wilddog}
    \tikzexternaldisable
    \tikzsetnextfilename{grid_HQ_wild2dog}
    \input{fig/grid}
    \begin{tikzpicture}
        \node[inner sep=0, fill=white] (G1) at (0, 0) {
            \gridcp{wild2dog}
                   {0, ..., 8}
                   {input/input, 
                    EGSDE1/\egsde, 
                    EGSDE2/\egsdeDG,
                    UVCGAN1/\uvcgan,
                    UVCGAN2wo/\thename}
        };
    \end{tikzpicture}
\end{table*}

\begin{table*}[p]
    \centering
    \caption{\textbf{Sample translations for \wildcat on \afhq.} 
             Since no benchmarking algorithms studied this task, 
             we only show the input and \thename's translation. }
    \label{fig:grid_HQ_wildcat}
    \tikzexternaldisable
    \tikzsetnextfilename{grid_HQ_wild2cat}
    \input{fig/grid}
    \begin{tikzpicture}
        \node[inner sep=0, fill=white] at (0, 0) {
            \begin{tikzpicture}
                \node[inner sep=0] (G1) at (0, 0) {
                    \gridcp{wild2cat}
                           {0, ..., 8}
                           {input/input, UVCGAN2wo/\thename}
                };
                \node[inner sep=0, anchor=north] (G2) at ([yshift=-.1 * \width]G1.south) {
                    \gridcp[9]{wild2cat}
                              {9, ..., 17}
                              {input/input, UVCGAN2wo/\thename}
                };   
            \end{tikzpicture}
        };
    \end{tikzpicture}
\end{table*}

\begin{table*}[p]
    \centering
    \caption{\textbf{Sample translations for \malefemale on \celebahq.}}
    \label{fig:grid_HQ_male2emale}
    \tikzexternaldisable
    \tikzsetnextfilename{grid_HQ_male2female}
    \input{fig/grid}
    \begin{tikzpicture}
        \node[inner sep=0, fill=white] (G1) at (0, 0) {
            \gridcp{male2female}
                 {9, ..., 17}
                 {input/input, 
                  EGSDE1/\egsde, 
                  EGSDE2/\egsdeDG,
                  UVCGAN1/\uvcgan,
                  UVCGAN2wo/\thename}
        };
    \end{tikzpicture}
\end{table*}

In this section, we provide additional translation samples to facilitate visual comparison of image quality. 
\autoref{fig:grid_LQ_selfieanime} and \ref{fig:grid_LQ_animeselfie} demonstrate samples on the \anime dataset.
\autoref{fig:grid_LQ_malefemale} and \ref{fig:grid_LQ_femalemale} provide gender swap samples on the \celeba dataset. 
\autoref{fig:grid_LQ_rmvGlasses} and \ref{fig:grid_LQ_addGlasses} show eyeglasses removal and addition samples on the \celeba dataset. 
Finally, \autoref{fig:grid_HQ_catdog}, \ref{fig:grid_HQ_wilddog}, and \ref{fig:grid_HQ_wildcat} provide samples on the \afhq dataset, and \autoref{fig:grid_HQ_male2emale}, \malefemale samples on the \celebahq dataset.

%% file: src/app_consistency_loss.tex
\section{Effects of the Pixel-Wise Consistency Loss}

The main part of the paper compares models trained with two settings of the pixel-wise consistency loss. The \thename model having $\lambda_\text{consist} = 0$ and \thename-C model with $\lambda_\text{consist} = 0.2$. In this section, we show an ablation of the $\lambda_\text{consist}$ values and their effect on the model performance.

\autoref{tab:app_results_lambda_consist} demonstrates the effect of different values of $\lambda_\text{consist}$ on the \thename model realism (as measured by the FID scores) and
pixel-wise faithfulness (as measured by the pixel-wise $L_2$ distance).

As one might expect, the increase in $\lambda_\text{consist}$ is accompanied by an improvement in pixel-wise image faithfulness and a decrease in image
realism. Values of $\lambda_\text{consist}$ below $0.2$ allow one to significantly improve pixel-wise image faithfulness at the expense of a modest loss of image realism. Further increases in $\lambda_\text{consist}$ produce
larger improvements in pixel-wise faithfulness, but also lead to significant decreases in image realism.

Additionally, \autoref{tab:app_results_lambda_consist} demonstrates that the trade-offs of image realism to pixel-wise faithfulness are not uniform across the datasets.
High values of $\lambda_\text{consist}$ allow one to achieve rather large improvements in pixel-wise faithfulness on the \malefemale task at a relatively
small loss of image realism. On the other hand, \catdog translation is subject
to a catastrophic loss of image realism with the increase of $\lambda_\text{consist}$.

%% file: src/app_uvcgan.tex
\section{\uvcgan Training Procedure}
\label{sec:app_train_uvcgan}

In this appendix, we provide details of the training setup of \uvcgan models for the
high-quality datasets (\celebahq and \afhq). The \uvcgan paper~\cite{torbunov2023uvcgan}
did not study these datasets, therefore we trained the corresponding models (\celebahq Male-to-Female, Cat-to-Dog, and Wildlife-to-Dog) from scratch.

Before training \uvcgan I2I translation generators, we performed their self-supervised
pre-training.
While performing the pre-training, we closely followed the original procedure~\cite{torbunov2023uvcgan}.
For reference, this pre-training procedure is the same procedure we used for the \thename generator pre-training.

At the second step, we trained the actual I2I translators, with the help of the
provided training scripts~\cite{uvcgan_software}. To closely match the \uvcgan paper, we performed a hyperparameter sweep over $\lambda_\text{cyc}$, $\lambda_\text{GP}$,
and $\gamma$. We explored $\lambda_{cyc}$ values of $[ 1, 5, 10 ]$. For the GP parameters $(\lambda_\text{GP}, \gamma)$, we explored the following configurations
$[ (10, 1), (1, 10), (0.1, 100), (1, 750) ]$, suggested by the \uvcgan source code
repository.

The best \uvcgan performance for all the high-quality translation directions
(Male-to-Female, Cat-to-Dog, Wildlife-to-Dog) is achieved for the same configuration of the tested hyperparameters: $\lambda_\text{cyc} = 5, \lambda_\text{GP} = 0.1, \gamma = 100$.
This configuration is used when presenting the \uvcgan results in the main part of the paper.